\documentclass{article}

\let\paperaddcontentsline\addcontentsline
\usepackage{iclr2027_conference,times}
\let\addcontentsline\paperaddcontentsline
\usepackage{amsmath,amsfonts,bm}

\def\eqref#1{equation~\ref{#1}}
\def\1{\bm{1}}

\DeclareMathAlphabet{\mathsfit}{\encodingdefault}{\sfdefault}{m}{sl}
\SetMathAlphabet{\mathsfit}{bold}{\encodingdefault}{\sfdefault}{bx}{n}

\DeclareRobustCommand{\proposedBench}{\texorpdfstring{CoM\textsuperscript{2}-Bench}{CoM2-Bench}}
\usepackage{graphicx}
\usepackage{booktabs}
\usepackage{pifont}
\usepackage{makecell}
\newcommand{\cmark}{\ding{51}}
\newcommand{\xmark}{\ding{55}}

\usepackage{float}
\usepackage{algorithmic}

\usepackage{wrapfig}
\usepackage{subcaption}
\usepackage{enumitem}
\usepackage{algorithm}
\usepackage[skins,breakable,listings]{tcolorbox}
\newtcblisting{promptbox}[1]{%
    enhanced,
    breakable,
    listing only,
    listing engine=listings,
    colback=black!5,
    colframe=black!50,
    colbacktitle=black!5,
    coltitle=black,
    arc=4pt,
    boxrule=0.5pt,
    boxsep=0pt,
    left=10pt,right=10pt,top=10pt,bottom=10pt,
    width=\linewidth,
    fonttitle=\bfseries,
    title={#1},
    listing options={%
        language={},
        basicstyle=\small\ttfamily,
        columns=fullflexible,
        keepspaces=true,
        showstringspaces=false,
        showspaces=false,
        showtabs=false,
        breaklines=true,
        breakatwhitespace=false,
        breakautoindent=false,
        breakindent=0pt,
        numbers=none,
        aboveskip=0pt,
        belowskip=0pt
    }
}
\usepackage{hyperref}

\usepackage{xspace}
\newcommand{\ie}{\textit{i.e.},\xspace}

\title{Grounding Confusable Movie Moments}

\author{\textbf{Dong-Hee Kim$^{1,*}$ \quad Seonwoo Choi$^{1,*}$ \quad Changbeen Kim$^{1}$ \quad Juyeon Ko$^{2}$ \quad Youngju Choi$^{2}$} \\
\textbf{Il Hyeon Mun$^{2}$ \quad Hyunwoo J. Kim$^{3}$ \quad Donghyun Kim$^{1}$} \\
$^{1}$Korea University \quad $^{2}$KT \quad $^{3}$KAIST \\
\texttt{\{dongheekim, doldol9080, d\_kim\}@korea.ac.kr}}

\iclrfinalcopy

\begin{document}
% Hide main body entries when the appendix contents file is read.
\addtocontents{toc}{\protect\setcounter{tocdepth}{-1}}

\maketitle
\lhead{Preprint}
{\let\thefootnote\relax\footnotetext{$^{*}$Equal contribution.}}

\vspace{2pt}
\begin{figure}[H]
    \centering
    \includegraphics[width=\linewidth]{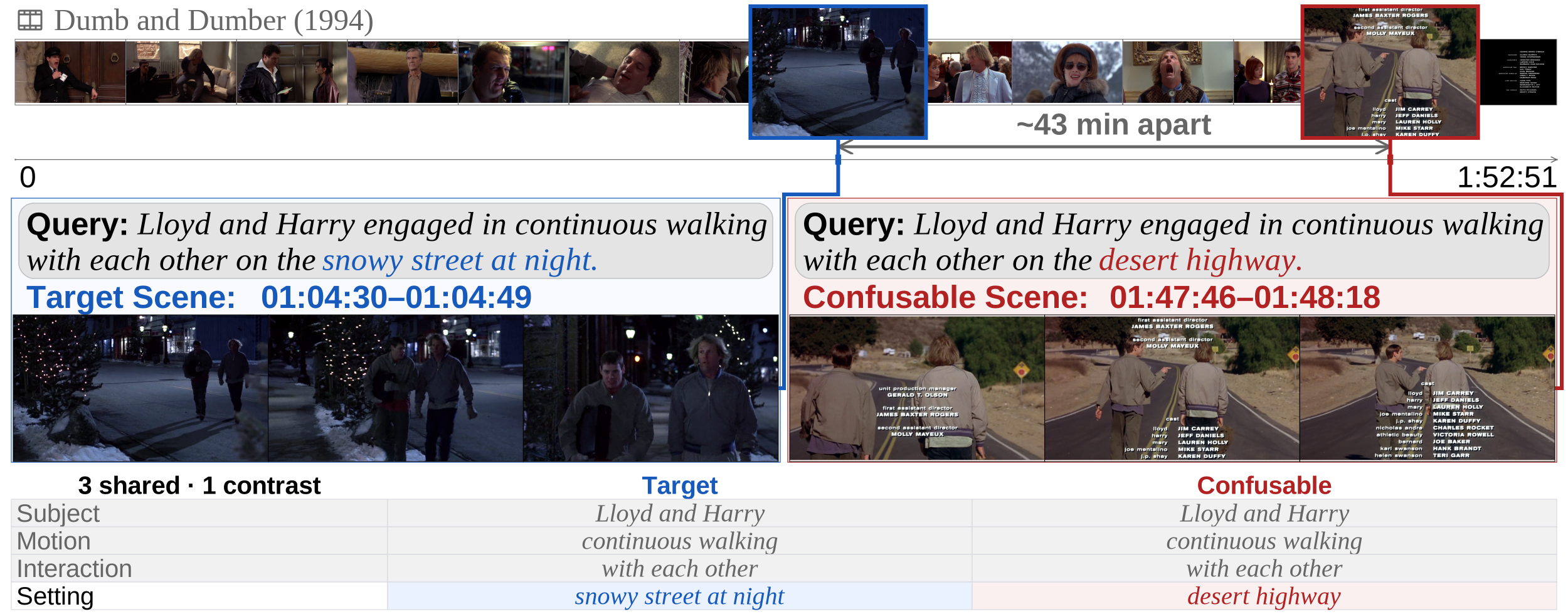}
    \caption{\textbf{An example of confusable moment grounding in a full-length movie from \proposedBench{}.} Each pair in \proposedBench{} consists of two moments whose queries share subject, motion, and interaction but differ only in setting: a \textcolor{blue}{snowy street} at night vs.\ a \textcolor{red}{desert highway}. Each query is grounded independently over the full movie, so success requires verifying not only the shared attributes that dominate query-moment similarity but also the distinguishing details.
    \proposedBench{} contains 46 movies totaling 90.4 hours, with 3,014 queries forming 1,507 confusable query pairs.}
    % The target and confusable intervals occur about 43 minutes apart in the same movie and share the same characters, motion, and interaction. 
    % Only the setting, a snowy street at night versus a desert highway in daytime, distinguishes the two queries and their ground-truth intervals. 
    \label{fig:overview}
\end{figure}
\vspace{2pt}

\begin{abstract}
Characters, locations, and actions recur throughout a movie, so many of its moments differ only in fine-grained details. Grounding such a moment in a full-length movie requires resolving that detail rather than matching the attributes it shares with other moments, yet existing temporal movie grounding benchmarks rarely test this.
We introduce \proposedBench{}, a benchmark of confusable query pairs, where the two queries in each pair share most attributes but differ in a single distinguishing detail within a full movie, posing a challenge for recent temporal grounding approaches.
% , and each query is grounded independently within a full movie.
%%
To locate where grounding breaks down, we analyze a retrieve-then-localize pipeline, in which a retriever first selects a short candidate chunk and a localizer then performs temporal grounding within it. Separating the two stages lets us attribute each failure to either retrieval or localization. 
% Retrieval is the bottleneck, and pairwise tests show the retriever separates a moment from its confusable partner far less reliably than from an unrelated one, so the difficulty is specific to confusable moments. 
% Its choice mostly follows the content the two moments share and misses the detail that tells them apart.
We find that retrieval is the major bottleneck: state-of-the-art retrievers often select a moment's confusable counterpart, because their choices follow the content the two moments share rather than the detail that tells them apart.
% Fine-tuning or reranking the retriever only partly closes this gap.
%%
% Because that missing detail must be read from the video, our intervention inspects the candidate windows directly rather than changing the retriever.
% A second failure stems from fixed-length chunking, which can cut a moment at chunk boundaries, so candidate chunks must be adjusted to cover the full event.
To address these failures, we propose a simple, training-free remedy, Agentic Window Search (AWS): an MLLM agent \textit{Decomposes} the query into attribute-level sub-queries, then \textit{Inspects} each candidate window against them and \textit{Expands} the window when evidence is cut off at its edge.
%%
% We evaluate AWS  corrected without retraining, and this inspection raises mIoU from 18.5 to 26.8 with the retriever and localizer fixed.
AWS works with existing retrievers and localizers without additional training, raising mIoU from 18.5 to 26.8.
\end{abstract}

\section{Introduction}
\label{sec:introduction}

%% Motivation
Movies repeatedly bring the same characters, locations, and actions back on screen, producing look-alike moments that differ only in fine-grained details. 
For example, in \emph{Dumb and Dumber} (Figure~\ref{fig:overview}), Lloyd and Harry walk together in two moments. Grounding one specific moment therefore requires another detail: whether they are on a snowy street at night or on a desert highway.
%% need to complement motivation
Video temporal grounding (VTG) localizes the start and end times of an event described by a natural-language query~\citep{gao2017tall}. As grounding extends from short clips to full-length movies, such repetition becomes the norm rather than the exception: the longer the video, the more moments tend to match most of a query. A grounding model for movie search, editing, or audio description must therefore find the one moment that satisfies every part of the query, not merely its most salient part.

%% limitation of existing bench
Existing benchmarks, however, cannot tell whether models do this. Their queries can often be localized to a single salient entity or action, so a model that ignores the remaining details can still score well, and its errors on confusable moments go unmeasured. Long-video benchmarks~\citep{soldan2022mad,yuan2025momentseeker} enlarge the search space, and compositional benchmarks~\citep{zou2026comet} combine temporal and spatial conditions, but neither controls whether another moment in the same video shares most of the query. Some construction pipelines even remove repeated events within each video~\citep{seo2026extremewhenbench}, discarding exactly the cases where this distinction matters. Although repeated descriptions within a movie are a recognized source of grounding ambiguity~\citep{soldan2022mad}, to our knowledge no benchmark deliberately constructs and verifies such cases. Current scores therefore cannot reveal whether a model grounds the full query or only the attributes it shares with other moments.

% In Dumb and Dumber, Lloyd and Harry walk together on a snowy street at night in one scene and along a desert highway in daytime in another, so the characters and action alone do not distinguish the moments (Figure~\ref{fig:overview}). 
% Video temporal grounding (VTG) locates a queried event's start and end times~\citep{gao2017tall}, which here requires checking the setting as well as the characters and action. Although repeated descriptions within a movie are a recognized source of potential grounding ambiguity~\citep{soldan2022mad}, coverage of long videos~\citep{soldan2022mad,yuan2025momentseeker} and compositional temporal and spatial conditions~\citep{zou2026comet} alone does not isolate whether models distinguish moments sharing most query attributes. This distinction is further obscured when repeated events are removed within each video during benchmark construction~\citep{seo2026extremewhenbench}, limiting the opportunity to test whether a model checks the identifying detail rather than relying on the attributes shared across moments. 

%% Our benchmark
We therefore study whether temporal grounding models distinguish paired \textit{confusable} movie moments that share most attributes but differ in one identifying detail, requiring them to match the query's combination of attributes. We build \proposedBench{} to answer this, with 3,014 confusable queries over the original videos of 46 movies. Its design makes correct grounding depend on the distinguishing detail. Each pair shares three attributes and differs in one, and the two queries are written in parallel so that only this detail, not wording or sentence structure, separates them. We ground each query on its own, without its partner as a contrast, so a model cannot lean on the comparison and must locate the detail in the video. Because a confusable pair is resolved only when both of its moments are grounded correctly, we introduce PairSuccess, a pair-level metric that counts a pair as correct only when both independently grounded queries exceed a given IoU threshold. Human annotators finally check each query for validity, temporal boundaries, and uniqueness under explicit guidelines, with blind cross-review and adjudication, keeping the labels reliable. \proposedBench{} is the only benchmark that pairs verified confusable moments in full-length movies (see Table~\ref{tab:benchmark_comparison}).

%% Analysis. 
On \proposedBench{}, we first ask where the state-of-the-art retrieve-then-localize pipeline fails, following prior work that narrows the temporal search space before precise localization~\citep{hannan2025revisionllm,seo2026extremewhenbench}. 
% explaining the framework
We adopt Movie RAG~\citep{kim2026narrascene} for this analysis, with a chunk retriever that ranks video chunks by embedding similarity to the query and a temporal localizer, hereafter the retriever and the localizer. We use TimeLens2~\citep{zhu2026timelens2} as the localizer, which predicts boundaries within a window, initially one retrieved chunk.
Our analysis identifies two limitations.
% limitation 1 -> motivation for query decomposition and matching
First, the retriever is the dominant bottleneck:
replacing the top-1 retrieved chunk with the \emph{oracle} chunk,
the one overlapping the ground truth most,
raises grounding sharply
(R@1@0.3 from 25.8 to 77.3, Table~\ref{tab:oracle_neighborhoods}),
so improving retrieval is the primary lever for performance.
We trace this low retrieval accuracy to the retriever's inability
to separate confusable pairs (Figure~\ref{fig:pairwise_confusion}).
% limitation 2 -> motivation for bounded expansion in AWS
Second, boundary errors persist even with the best overlapping chunk,
partly because fixed chunks cut events at their edges,
and extending that chunk within a small budget raises R@1@0.7 from 34.8 to 43.2 (Table~\ref{tab:oracle_neighborhoods}).

% % analysis
% Our analysis identifies window retrieval as the main bottleneck, consistent with an earlier failure analysis attributing most sampled errors of a Video-LLM on hour-scale videos to search~\citep{seo2026extremewhenbench}.
% Supplying the oracle chunk, the chunk that overlaps the ground truth most, improves grounding substantially more than perfect boundary prediction within the retrieved window could, assuming predictions remain inside that window. 
% %
% Yet boundary errors persist even with the best overlapping chunk. Fine tuning the retriever~\citep{oord2018infonce} and pretrained reranker~\citep{li2026qwen3} improve grounding but leave a substantial gap to the gold chunk reference. Moreover, pairwise comparisons show that the retriever separates the correct chunk from its confusable pair less reliably than from a random chunk, exposing the limits of similarity alone. A query reduced to only the attributes the pair shares already decides most of these choices, and restoring the distinguishing detail moves the retriever to the correct chunk in fewer than half of the cases where it should.

%% AWS our method
To address these failures, we propose training-free Agentic Window Search (AWS), which operates within the retrieve-then-localize pipeline: between the two stages, an agent verifies the candidate window directly against the query or adjusts its length before passing one to the localizer.
% To test whether this failure can be corrected without retraining, we examine candidate evidence directly, an intervention we call Agentic Window Search (AWS). 
AWS first \textit{Decomposes} the query into attribute-level sub-queries, so that the distinguishing detail is checked on its own rather than absorbed into the shared content. It then \textit{Inspects} the top-ranked candidate windows in retrieval order, accepting a window only when every sub-query holds, and \textit{Expands} a window within a fixed budget when evidence is cut off at its edge.
 % The agent decomposes the query into sub-queries and checks each against windows initialized from ranked chunks, using relevant character references to accept or reject candidates. When evidence continues beyond a window boundary, it extends the window within a fixed budget. The unchanged localizer receives either an accepted window or, if none is accepted, one selected by a fallback comparison. The search thus checks evidence before boundary prediction without task-specific training.
% Building on candidate verification~\citep{liu2026videomind}, verified narrative memory across scenes~\citep{hahm2026storyteller}, and visual character references for naming in audio description without task-specific training~\citep{xie2024autoadzero}, we use character memory to guide window selection and expansion while keeping the retriever and localizer fixed. This memory links names to reference images from dialogue and nearby frames without evaluation queries or ground truth, helping identify characters without imposing appearance requirements absent from the query.
In addition, movie queries often name characters, but a name alone does not reveal who appears on screen. We therefore build a character memory that links names to reference frames using the movie's dialogue and nearby frames, without evaluation queries or ground truth. The agent uses these references only to identify named characters during inspection, without adding appearance constraints absent from the query.

Incorporated into the existing retrieve-then-localize pipeline of Movie RAG, AWS raises mIoU from 18.5 to 26.8. These gains show that verifying candidate evidence improves grounding without additional training, while the remaining gap to the oracle leaves room to improve both window selection and boundary prediction.
% We discuss related benchmarks and methods in Appendix~\ref{sec:related-work}.
Our contributions are summarized below.
\begin{itemize}[leftmargin=1.5pc, itemsep=2pt, topsep=2pt]
    \item We introduce \proposedBench{}, a full-movie temporal grounding benchmark of human-verified confusable query pairs, in which the two queries of each pair share most attributes but differ in a single distinguishing detail.
    % We introduce \proposedBench{}, a full movie temporal grounding benchmark with confusable query pairs.
    \item We identify chunk retrieval as the main bottleneck and show that the retriever fails to account for all attributes of a multi-attribute query: its choices are dominated by the attributes a moment shares with its confusable counterpart, while the single attribute that distinguishes them rarely changes the retrieved chunk.
    % We identify window retrieval as the main bottleneck and show that the retriever confuses paired moments because its choices follow shared content rather than the detail that separates them.
    \item We propose Agentic Window Search (AWS), a training-free, plug-in agent that verifies candidate windows against attribute-level sub-queries between retrieval and localization, raising mIoU from 18.5 to 26.8 without modifying or retraining the retriever or localizer.
    % We show this failure is actionable, since verifying candidate evidence directly between retrieval and localization (Agentic Window Search) substantially improves grounding without modifying the retriever or localizer.
\end{itemize}

\begin{table}[t]
    \centering
    \caption{\textbf{Comparison with existing video temporal grounding benchmarks.}
    \textit{Confusable moments}: other moments in the same video that match most of the query
    (\cmark\ deliberately constructed and verified, $\triangle$ present but uncontrolled).
    \textit{Attribute contrast}: paired queries differ in exactly one attribute with parallel phrasing.
    \textit{Pair metric}: success requires both queries of a pair to be grounded.
    \proposedBench{} is the only benchmark in which a model must separate a target from a verified confusable moment in the same full-length movie.}
    \vspace{-2mm}
    \label{tab:benchmark_comparison}
    \resizebox{\linewidth}{!}{%
    \begin{tabular}{lccccccc}
        \toprule
        Benchmark & Scope (video len., m) & Avg.\ target len.\ (s) & \# Queries & Confusable moments & Attribute contrast & Pair metric & Video access \\
        \midrule
        Charades-STA~\citep{gao2017tall}          & clip (0.51)       & 8.1  & 3,720  & \xmark      & \xmark & \xmark & \cmark \\
        QVHighlights~\citep{lei2021qvhighlights}  & clip (2.50)       & 24.6 & 1,542  & \xmark      & \xmark & \xmark & \cmark \\
        MomentSeeker~\citep{yuan2025momentseeker} & long video (20.03) & 12.8 & 1,800  & \xmark      & \xmark & \xmark & \cmark \\
        CoMET-Bench~\citep{zou2026comet}          & long video (33.81) & 6.9  & 2,789  & \xmark      & \xmark & \xmark & \cmark \\
        MAD~\citep{soldan2022mad}                 & full movie (110.77) & 4.1  & 72,000 & $\triangle$ & \xmark & \xmark & \xmark \\
        \midrule
        \textbf{\proposedBench{} (Ours)}          & full movie (117.8) & 60.2 & 3,014  & \cmark      & \cmark & \cmark & \cmark \\
        \bottomrule
    \end{tabular}%
    }
    \vspace{-2mm}
\end{table}

\section{\proposedBench{}: Confusable Movie Temporal Grounding}
\label{sec:benchmark}

We introduce \proposedBench{} (Confusable Movie Moments-Bench), a full-movie temporal grounding benchmark designed to evaluate whether a model can distinguish semantically similar moments occurring at different times in the same movie. We summarize the differences from existing benchmarks in Table~\ref{tab:benchmark_comparison} and defer the full discussion of related work to Appendix~\ref{sec:related-work}.

\vspace{-5pt}
\paragraph{Task and Benchmark Design.}
Unlike conventional benchmarks in which a query can often be localized using a single salient entity or action, \proposedBench{} organizes queries into \emph{confusable query pairs}. Each pair refers to two temporally distant moments that share semantic attributes but differ in one attribute (\ie confusable moments).

Given a full movie $V$ of duration $T$ and a text query $q$, a model predicts an interval $\hat{\tau}=[s,e]$, where $0\leq s<e\leq T$. Each pair comprises a target query $q_t$ and a paired confusable query $q_c$, with ground truth intervals $\tau_t$ and $\tau_c$ in the same movie. Both queries are evaluated independently as targets without the paired query as input, while the other moment serves as a confusable alternative.
For the $i$-th of $N$ pairs,
we write its two queries as $q_{i,t}$ and $q_{i,c}$,
with ground-truth intervals $\tau_{i,t}$ and $\tau_{i,c}$
and predictions $\hat{\tau}_{i,t}$ and $\hat{\tau}_{i,c}$.
% For $N$ confusable pairs, let $\hat{\tau}_{i,j}$ and $\tau_{i,j}$ denote the predicted and ground truth intervals for query $j\in\{t,c\}$ in pair $i$.
We define $\operatorname{PairSuccess}@\theta=(100/N)\sum_{i=1}^{N}\prod_{j\in\{t,c\}}\mathbf{1}[\operatorname{IoU}(\hat{\tau}_{i,j},\tau_{i,j})\geq\theta]$, the percentage of confusable pairs whose two queries are both grounded with temporal IoU of at least $\theta$.
Missing or invalid predictions count as failures, and all $N$ pairs remain in the denominator.
We call the ground-truth interval of a query a moment. During construction, moments originate from events that an MLLM proposes within subtitle-anchored clips (Appendix~\ref{sec:app_construction}).
Pair identity is used only for aggregation, and the paired query, contrast label, and ground truth are never provided as inference input.
Paired moments share three semantic attributes and differ in one, so that success requires resolving the discriminative attribute. Attributes include participating characters, action, object, setting, or temporal relation. Both queries use the same attribute roles and parallel linguistic structure. Figure~\ref{fig:overview} illustrates events sharing characters, motion, and interaction but differing in setting. This construction tests sensitivity to the distinguishing attribute without guaranteeing that every query requires all four attributes to be resolved jointly.

\vspace{-5pt}
\paragraph{Dataset Construction.}
We source movies and subtitles from ALLVB~\citep{tan2025allvb}.
Subtitle-anchored clip segmentation groups consecutive cues into clips and expands their visual context using adjacent clip boundaries.
A multimodal large language model (MLLM) uses the expanded video and subtitle derived clip descriptions to generate event intervals and structured attributes.
We construct parallel confusable queries from shared and contrasting attributes, retaining pairs only after rigorous human verification.
Details about construction and human verification are in Appendix~\ref{sec:app_benchmark}.

\begin{figure}[t]
    \vspace{-5pt}
    \centering
    \begin{subfigure}[t]{0.24\linewidth}
        \centering
        \includegraphics[width=\linewidth]{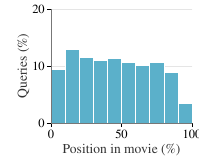}
        \caption{GT position}
    \end{subfigure}\hfill
    \begin{subfigure}[t]{0.24\linewidth}
        \centering
        \includegraphics[width=\linewidth]{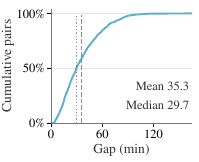}
        \caption{Temporal distance}
    \end{subfigure}\hfill
    \begin{subfigure}[t]{0.24\linewidth}
        \centering
        \includegraphics[width=\linewidth]{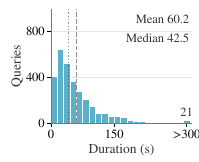}
        \caption{Moment duration}
    \end{subfigure}\hfill
    \begin{subfigure}[t]{0.24\linewidth}
        \centering
        \includegraphics[width=\linewidth]{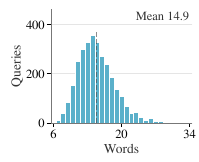}
        \caption{Query length}
    \end{subfigure}
    \vspace{-2mm}
    \caption{\textbf{Benchmark statistics.} Distributions of ground truth positions within movies, temporal distance between paired moments, durations of moments, and query lengths. Positions use interval midpoints normalized by movie duration, dashed and dotted lines mark means and medians, and the duration overflow bin includes all intervals longer than 300\,s.}
    \label{fig:benchmark_stats}
    \vspace{-10pt}
\end{figure}

\vspace{-5pt}
\paragraph{Benchmark Statistics.}
The current verified release of \proposedBench{} contains 46 full-length movies with a combined duration of 90.4 hours. The average movie duration is 117.8 minutes.  Human verification retains 1,507 pairs, corresponding to 3,014 temporal-grounding queries. The average query length is 14.9 words. Ground-truth moments have an average duration of 60.2\,s and a median duration of 42.5\,s. The paired moments are separated by an average of 35.3 minutes, with a median of 29.7 minutes, measured from the end of the earlier ground truth interval to the start of the later one. These statistics show that the benchmark emphasizes temporally distant confusable moments rather than local boundary distinctions.
Figure~\ref{fig:benchmark_stats} summarizes the distributions of ground truth positions, temporal distance of paired movie moments, moment durations, and query lengths.

Ground-truth moments are spread widely across movies, reinforcing the need for full movie search. With interval midpoints normalized by movie duration, each of the first nine deciles contains roughly 9 to 13\% of queries, while the final decile, which largely covers end credits, contains 3.5\%.

\vspace{-5pt}
\paragraph{Comparison with Existing Benchmarks.}
Table~\ref{tab:benchmark_comparison} compares \proposedBench{} with existing benchmarks in scale and design. \proposedBench{} is the only benchmark that deliberately constructs
and verifies confusable moments within the same full movie.
MAD releases only precomputed CLIP features rather than the videos,
and its average target length is only 4.1\,s.
In contrast, \proposedBench{} builds on the movie release of
ALLVB~\citep{tan2025allvb}, enabling evaluation on raw frames,
and its targets are complete moments averaging 60.2\,s.
\section{Diagnosing Confusable Movie Moment Temporal Grounding}
\label{sec:diagnosis}

% If the section starts near a page bottom, the wrap may overflow. Move the opening paragraph and wrap together to a page with enough space.

With \proposedBench{} in place, we ask how current grounding methods handle it and where they break down.
We analyze a retrieve-then-localize pipeline, whose separate retrieval and localization stages let us attribute a failure to one stage, and adopt Movie RAG~\citep{kim2026narrascene} as a representative and competitive instance (Table~\ref{tab:frozen_public_models}).
Movie RAG has a chunk retriever that ranks movie chunks by query similarity and a temporal localizer that predicts event boundaries within the supplied window.
We refer to these components as the retriever and the localizer.

\begin{wrapfigure}{r}{0.44\linewidth}
    \vspace{-40pt}
    \centering
    \includegraphics[width=\linewidth]{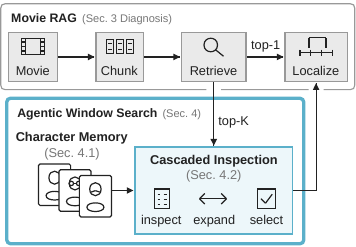}
    \caption{\textbf{Movie RAG and Agentic Window Search.} The upper row shows the retrieve-then-localize pipeline diagnosed in Section~\ref{sec:diagnosis}. The lower block replaces the window from the top-ranked chunk with agentic search over the top $K$ candidates described in Section~\ref{sec:method}.}
    \label{fig:method_overview}
    \vspace{-25pt}
\end{wrapfigure}

A chunk is a retrieval unit, while a window is the span supplied to the localizer, initially one retrieved chunk and possibly extended beyond its edges (Section~\ref{sec:agentic_verification}).
Movie RAG reports fixed-length chunks to be competitive with scene segmentation, so we partition movies into fixed 180\,s chunks for our diagnosis.
We instantiate the retriever with Qwen3-VL-Embedding 2B~\citep{li2026qwen3} and the localizer with TimeLens2 8B~\citep{zhu2026timelens2}, and use both throughout our diagnosis unless stated otherwise.
The retriever embeds each chunk using all frames sampled at 1 fps together with its subtitles, embeds the text query, and ranks chunks by cosine similarity between the chunk and query embeddings.
The upper part of Figure~\ref{fig:method_overview} shows this retrieve-then-localize pipeline.
We hold one stage fixed while replacing the other with an oracle to assess how window selection and boundary prediction limit performance.
All analyses in this section evaluate the full \proposedBench{}.

\begin{table}[t]
    \centering
    \small
    \caption{\textbf{Chunk retrieval, not localization, limits grounding.} Giving the localizer the oracle chunk (row 6) raises mIoU much more than a perfect localizer on the retrieved chunk (row 3, $\dagger$ marks upper bounds assuming predictions stay within the window). Fine-tuning the retriever or reranking (rows 4 and 5) closes only part of that gap. Extending the oracle chunk within the search's expansion budget (row 7) adds a further gain. The gray row 8 combines both oracles.}
    \vspace{-5pt}
    \label{tab:oracle_neighborhoods}
    \setlength{\tabcolsep}{3pt}
    \begin{tabular}{cllcccc}
        \toprule
        Row & Window selection & Localizer
        & R@1@0.3 & R@1@0.5 & R@1@0.7 & mIoU \\
        \midrule
        \multicolumn{7}{l}{\footnotesize\textit{Temporal grounding on full-movies}} \\
        1 & \xmark~(Full-movie) & TimeLens2
        & 10.0 & 6.9 & 3.9 & 7.4 \\
        \midrule
        \multicolumn{7}{l}{\footnotesize\textit{Base pipeline and bounds}} \\
        2 & Retrieved top-1 & TimeLens2
        & 25.8 & 19.1 & 12.2 & 18.5 \\
        3 & Retrieved top-1 & Oracle$^{\dagger}$
        & 34.8 & 31.9 & 28.1 & 31.0 \\
        \midrule
        \multicolumn{7}{l}{\footnotesize\textit{Retrieval fixes (Section~\ref{sec:diagnosis_retrieval_fixes})}} \\
        4 & Fine tuned retriever & TimeLens2
        & 32.4 & 24.1 & 15.3 & 23.1 \\
        5 & Top-5 reranking & TimeLens2
        & 33.4 & 25.1 & 16.1 & 24.3 \\
        \midrule
        \multicolumn{7}{l}{\footnotesize\textit{Oracle retriever reference}} \\
        6 & Oracle Top-1 & TimeLens2
        & 77.3 & 57.9 & 34.8 & 53.8 \\
        7 & Oracle Top-1 + Oracle expand & TimeLens2
        & 79.9 & 63.0 & 43.2 & 58.3 \\
        \textcolor{gray}{8} & \textcolor{gray}{Oracle Top-1} & \textcolor{gray}{Oracle$^{\dagger}$}
        & \textcolor{gray}{99.9} & \textcolor{gray}{99.6} & \textcolor{gray}{85.9} & \textcolor{gray}{91.6} \\
        \bottomrule
    \end{tabular}
    \vspace{-5pt}
\end{table}

\subsection{Where Does the Pipeline Fail?}
\label{sec:diagnosis_pipeline}

% Window retrieval is the dominant bottleneck in this pipeline, as shown by the oracle comparisons in Table~\ref{tab:oracle_neighborhoods}.
To identify the main cause of failure in the Movie RAG,
we vary the window selection and the localizer with oracle substitutes
and report the results in Table~\ref{tab:oracle_neighborhoods}.

\vspace{-5pt}
\paragraph{The retriever matters more than the localizer.} To assess how much window selection limits accuracy,
we first substitute the retriever's top-1 chunk
with an oracle top-1 chunk, the one that overlaps the ground-truth interval most.
% To assess how much window selection limits accuracy, we define the gold chunk as the 180-second chunk with maximal overlap with the ground truth interval, which need not contain the whole event.
Supplying this oracle chunk lets TimeLens2 reach 53.8 mIoU (Table~\ref{tab:oracle_neighborhoods}, row~6), far above the 31.0 mIoU that even an oracle localizer reaches on the retrieved top-1 chunk (row~3).
These results indicate frequent retrieval misses, with the retrieved top-1 chunk having no overlap with the ground truth for 62.8\% of queries, as well as incomplete coverage when overlap exists.
Specifically, for a single continuous ground truth interval, the maximal attainable IoU equals the fraction covered by the retrieved window, assuming the predicted interval lies entirely within that window.
The same ceiling limits even the joint oracle in row~8 to 91.6 mIoU.
Window selection thus caps what any localizer can achieve, since a localizer can only place boundaries inside the window it receives. A better localizer therefore helps little while the window misses or cuts the event, and its gains grow only once retrieval supplies the right chunk.

\vspace{-5pt}
\paragraph{Fixed 180\,s chunking truncates events.}
Even with the oracle chunk, localization loses accuracy, partly because fixed chunks ignore event boundaries and can cut an event at their edges.
Extending the oracle chunk toward the event within the budget of Section~\ref{sec:method} (at most three 30\,s steps) raises mIoU from 53.8 to 58.3 (row~7), mostly at the strictest threshold (R@1@0.7 from 34.8 to 43.2), and the remaining gap to the oracle localizer reflects the localizer itself.
Growing a window past a fixed chunk edge thus recovers part of the loss, which motivates the bounded expansion in Section~\ref{sec:agentic_verification}.

\begin{figure}[t]
    \centering
    \begin{subfigure}[c]{0.34\linewidth}
        \centering
        \includegraphics[width=\linewidth]{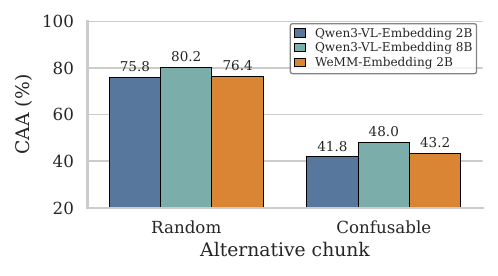}
        \caption{}
        \label{fig:pairwise_caa}
    \end{subfigure}\hfill
    \begin{subfigure}[c]{0.64\linewidth}
        \centering
        \includegraphics[width=\linewidth]{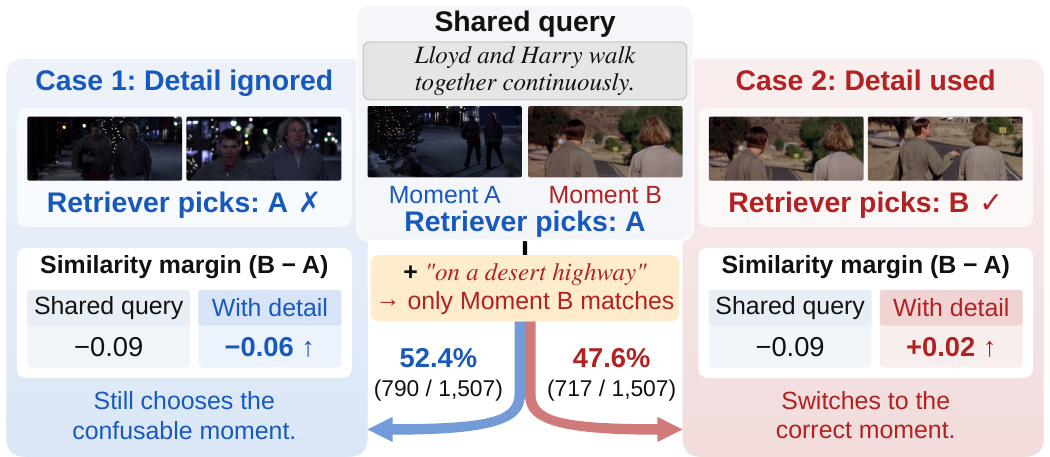}
        \caption{}
        \label{fig:shared_query}
    \end{subfigure}
    \vspace{-7pt}
    \caption{\textbf{The retriever confuses paired moments because it follows shared content.} (a) Chance-adjusted accuracy (CAA) at picking the correct chunk over one alternative, where 0\% is chance. Every retriever loses much of its discrimination against the paired confusable moment. (b) For each query, we drop the distinguishing detail to form a shared query and compare the retriever's margin between the two paired oracle chunks with and without the detail. Among the queries from \proposedBench{}, adding the detail switches only 47.6\% to the correct one.}
    \label{fig:pairwise_confusion}
    \vspace{-10pt}
\end{figure}

\subsection{Do confusable moments cause the failure?}
\label{sec:diagnosis_bottleneck}
Since retrieval remains the dominant bottleneck, we next ask whether the retriever can tell oracle chunks from their paired confusable alternatives.
To isolate this, we compare each query's oracle chunk against one alternative and select the chunk with higher query similarity under the retriever.
In the Confusable condition, the alternative is the paired confusable query's oracle chunk.
In the Random condition, accuracy is the exact expectation over uniformly sampled chunks from the same movie after excluding the oracle chunks of both paired queries and any chunk overlapping either ground truth interval.
Reporting chance-adjusted accuracy (CAA), where 0\% is chance and 100\% is always correct (Figure~\ref{fig:pairwise_caa}), the default retriever keeps 75.8\% of its discrimination against a random chunk but only 41.8\% against the confusable one, about half.
A larger Qwen3-VL-Embedding 8B and a different embedding model, WeMM-Embedding~\citep{wemm-embedding}, show the same drop, so the confusion reflects neither model scale nor one particular embedding model.
This raises the question of whether the retriever uses the detail that separates a pair at all.

% Full-movie retrieval remains substantially harder, with R@1 of 29.8\%.
% The two alternative conditions each test discrimination against a single chunk, while retrieval R@1 requires the gold chunk to outrank every candidate in the movie. A gold chunk can therefore beat most individual random alternatives yet lose to a higher scoring candidate in the full pool.

% TODO restore once the removal experiment (TimeLens2 on changed queries) is run
% The effect of confusable chunks on final grounding performance remains to be established.
% To measure this effect, we remove the confusable chunk from the retrieval pool while holding the query and localizer fixed, and compare the resulting selection and grounding performance with the original pool and with removal of a random or matched hard negative.
% Table~\ref{tab:confusable_removal} provides the comparison, with results pending.
% 
% \begin{table}[t]
%     \centering
%     \caption{Effect of removing one candidate from the retrieval pool while holding the query and localizer fixed.}
%     \label{tab:confusable_removal}
%     \resizebox{\linewidth}{!}{%
%     \begin{tabular}{lccccc}
%         \toprule
%         Candidate pool & Top-1 changed (\%) & Gold retrieval R@1 & mIoU & $\Delta$mIoU & R@1@0.5 \\
%         \midrule
%         Original & 0 & -- & -- & 0 & -- \\
%         Remove random negative & -- & -- & -- & -- & -- \\
%         Remove matched hard negative & -- & -- & -- & -- & -- \\
%         Remove confusable chunk & -- & -- & -- & -- & -- \\
%         \bottomrule
%     \end{tabular}%
%     }
% \end{table}

\subsection{Does the Retriever Use the Distinguishing Detail?}
\label{sec:diagnosis_shared_query}

The retriever's success on confusable pairs comes mostly from the content the two moments share, and the distinguishing detail is decisive in fewer than half of the cases where it is needed.
For each pair, we prompt an LLM to write a shared query from the attributes common to the two queries, dropping the detail, and we compare it with each original query under the same retriever and pairs as above.
For example, the shared query ``Lloyd and Harry walk together continuously.'' matches both moments of Figure~\ref{fig:overview} (Figure~\ref{fig:shared_query}), the snowy street and the desert highway, and only restoring the setting should send each query to its correct moment.
A query prefers its correct moment when its similarity to its own oracle chunk exceeds that to the confusable moment's oracle chunk, and we call this difference the margin.
Since the shared query fits both moments, its preference says nothing about the correct one, and an ideal retriever would let the detail move every original query to its correct moment.
Instead, when the shared query prefers the confusable moment, the detail moves only 47.6\% of queries to the correct one (Figure~\ref{fig:shared_query}), and when it already prefers the correct moment, the detail leaves 94.2\% of choices unchanged, with the full distribution in Appendix~\ref{sec:app_retriever}.
The drop in Section~\ref{sec:diagnosis_bottleneck} thus comes from similarity following the shared content, so recovering the detail calls for checking it against the video rather than reweighting similarity.

\subsection{Can Simple Retrieval Fixes Close the Gap?}
\label{sec:diagnosis_retrieval_fixes}

\paragraph{Fine-tuning the retriever with auto-generated training data.}
% Fine-tuning the retriever and pretrained reranker both help with TimeLens2 fixed, yet neither closes the gap to the oracle retriever.
The first fix fine-tunes Qwen3-VL-Embedding 2B on 5,730 queries in 2,865 confusable pairs, auto-generated without human review by our benchmark pipeline (Appendix~\ref{sec:app_construction}) from ALLVB 64 movies outside the \proposedBench{}.
Each pair supplies a hard negative, so for every query we pair its target moment's oracle chunk as the positive with the confusable moment's oracle chunk as the negative, and a contrastive InfoNCE objective~\citep{oord2018infonce} with in-batch negatives trains the retriever to prefer the target chunk.

\vspace{-5pt}
\paragraph{Reranking.} The second fix leaves the retriever alone and adds the Qwen3-VL Reranker~\citep{li2026qwen3}.
Instead of scoring a chunk and query from separate precomputed vectors, it reads the two together and returns one relevance score for each of the top five chunks, the same candidates our search later inspects.
These fixes, rows~4 and~5 of Table~\ref{tab:oracle_neighborhoods}, lift mIoU from 18.5 to 23.1 and 24.3, still far below the 53.8 the same localizer reaches on the oracle chunk.
Both fixes still score the query as a whole, so a candidate matching the shared content can outrank one matching the detail. This calls for checking each part of the query against the candidate, which we take up in Section~\ref{sec:method}.
Section~\ref{sec:analysis} then evaluates whether combining either fix with this verification yields further gains.

\section{Direct Evidence Verification by Agentic Window Search}
\label{sec:method}

\begin{figure}[t]
    \centering
    \includegraphics[width=\linewidth]{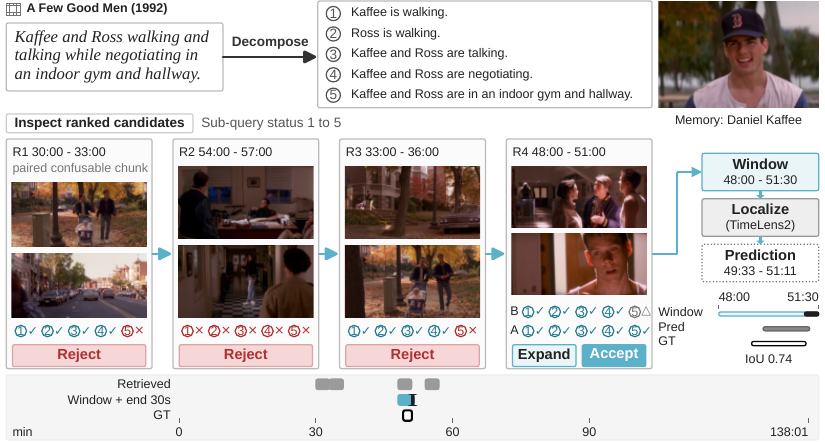}
    \vspace{-14pt}
    \caption{\textbf{Example of Agentic Window Search on the movie ``A Few Good Men''.} \textsc{Decompose} forms numbered sub-queries, and \textsc{Inspect} checks each against candidate windows with a character reference, rejecting candidates that miss a sub-query, and \textsc{Expand} adjusts a window whose evidence is cut off at its edge before accepting it.}
    \label{fig:qualitative}
    \vspace{-11pt}
\end{figure}

The retriever folds every attribute of a query into one score, so its choice follows what a pair shares (Section~\ref{sec:diagnosis_shared_query}). Agentic Window Search (Figure~\ref{fig:qualitative}) instead splits the query into sub-queries and checks each against candidate windows.
It inspects candidates in retrieval order, extending windows within a fixed budget (Section~\ref{sec:agentic_verification}), and links named characters to visual references (Section~\ref{sec:character_memory}).
Unlike in Movie RAG, the returned window $W$ may come from any rank and extend past chunk edges, and the unchanged localizer predicts $\hat{\tau}=[s,e]$ from $q$ and the video restricted to $W$, keeping the retriever and localizer fixed with no task-specific training.

\subsection{Cascaded Inspection with Bounded Expansion}
\label{sec:agentic_verification}
Inspection checks every part of the query, not its overall similarity.
A single MLLM performs \textsc{Decompose}, \textsc{Inspect}, and \textsc{Select} through zero-shot prompting without task-specific training.
Given query $q$, ranked chunks $C_{1:K}$, and character memory (Section~\ref{sec:character_memory}), \textsc{MatchCharacters} produces candidates $P$ by name lookup in memory without a model call.
A single \textsc{Decompose} call splits $q$ into a checklist $D$ of sub-queries, so that the distinguishing detail is checked on its own rather than absorbed into the shared content, and selects characters from $P$ with one reference each to form $R$, keeping both fixed across candidates.
Because fixed chunks can cut an event at their edges (Section~\ref{sec:diagnosis_pipeline}), candidate inspection distinguishes missing evidence from evidence cut off at a window edge.
\textsc{Inspect} evaluates the sampled frames and subtitles of the candidate window with the references in $R$ and assigns each sub-query a status of true, false, or cutoff.
It returns \textsc{Accept} when every sub-query holds somewhere in the window, \textsc{Reject} for missing evidence, or \textsc{Expand} for evidence cut off at an edge.
The same call chooses the expansion side and duration, and \textsc{ExpandWindow} applies it without a model call, clamps the window to movie boundaries, and appends frames and subtitles from the added span.
The budget permits at most $M$ expansions per candidate, each up to $\Delta_{\max}$ seconds.
Once the budget is exhausted, \textsc{Expand} is disabled, and any residual \textsc{Expand} response is converted to \textsc{Reject}. The full pseudocode is in Appendix~\ref{sec:app_cascaded}.
% , implemented as EMIT,

Candidates are inspected in retrieval order, and the loop stops at the first \textsc{Accept}, so a high-ranked candidate that satisfies the query ends the search early.
If all candidates are rejected, a single \textsc{Select} call compares their final summaries, rejection confidences, and midpoint frames and returns the selected window.
The procedure thus returns a window $W$ for the localizer with bounded model calls.
Excluding retries of malformed outputs, the worst case requires $K(M+1)$ \textsc{Inspect} calls, one \textsc{Decompose} call, and at most one \textsc{Select} call.

\subsection{Character Memory Construction}
\label{sec:character_memory}
Sub-queries often name characters, such as Lloyd and Harry in Figure~\ref{fig:overview}, and \textsc{Inspect} can verify them only if it knows each character's appearance, which neither the query nor the window states. We therefore build character memory from movie video, subtitles, and the movie identifier, without evaluation queries or ground truth. Dialogue anchors identify names, nicknames, and forms of address through introductions, speaker identification, and direct address to participants, excluding possessive references and third-party mentions. Anchor timestamps select nearby frames for reference clips where the model links names to visible people. Comparisons use stable physical cues to merge names for the same character while preserving distinct characters and appearance variants. Memory stores names and reference frames checked for identity and quality across multiple clips. \textsc{Inspect} (Section~\ref{sec:agentic_verification}) uses these references to match a named character to a person on screen, without adding appearance constraints absent from the query or proving presence in a candidate window. Appendix~\ref{sec:app_memory} gives construction details.

\section{Experiments}
\label{sec:experiments}

\vspace{-5pt}
\paragraph{Setup.}
We instantiate the pipeline introduced in Section~\ref{sec:diagnosis} with the \texttt{Qwen3-VL-Embedding 2B} and TimeLens2.
Retrieval and inspection sample frames at 1 fps. Retrieval uses all sampled frames per chunk, together with subtitles.
We use an MLLM in AWS with \texttt{Qwen3.5-4b} and \texttt{Qwen3.6-27b}, using $K=5$ ranked candidates.
Verification permits $M=3$ expansions per candidate with $\Delta_{\max}=30$\,s per expansion.
Expansion uses the same sampling rate, with at most 32 frames per expansion.
Decompose and Select use temperature 0.2, while Inspect uses temperature 0.4 and top probability 0.9.
Thinking is disabled for all three roles.
Memory variants use the same stored character memory, while variants without memory give the verifier no character references.
% TODO Link the localizer's native sampling and exact checkpoint to the final predictions.
% Hardware and measured inference cost remain to be confirmed in Appendix~\ref{sec:app_cost}.

\vspace{-5pt}
\paragraph{Evaluation.}
AWS comparison keeps retrieval and localization fixed across verifier configurations.
Temporal IoU is $|\hat{\tau}\cap\tau|/|\hat{\tau}\cup\tau|$, where $\hat{\tau}$ and $\tau$ are the predicted and ground truth intervals and $|\cdot|$ denotes temporal duration.
We report mIoU as the mean temporal IoU over queries and R@1@$\theta$ as the fraction of queries whose single prediction achieves an IoU of at least $\theta$, with $\theta\in\{0.3,0.5,0.7\}$.
Both metrics are reported as percentages.
Queries with missing or invalid predictions receive an IoU of zero.
We also report PairSuccess@$\theta$ defined in Section~\ref{sec:benchmark} at the same thresholds.

% TODO specify public model input adaptations.
% TODO specify numerical precision and inferen0ce batching.
% TODO specify the accounting of retries, cache reuse, and offline memory construction in reported costs.

\begin{table}[t]
    \centering
    \caption{\textbf{Direct evidence verification improves grounding and pair success without training.} Public grounding models score low on \proposedBench{}, while Movie RAG is the strongest baseline. Checking candidate evidence raises every metric. P@$\theta$ is PairSuccess@$\theta$, the percentage of confusable pairs whose two queries are both grounded with IoU at least $\theta$.} % Other retriever and localizer combinations are in Appendix~\ref{sec:app_pipeline_combos}.
    \vspace{-5pt}
    \label{tab:frozen_public_models}
    \setlength{\tabcolsep}{3pt}
    \resizebox{0.9\linewidth}{!}{%
    \begin{tabular}{lccccccc}
        \toprule
        Model & P@0.3 $\uparrow$ & P@0.5 $\uparrow$ & P@0.7 $\uparrow$ & R@1@0.3 $\uparrow$ & R@1@0.5 $\uparrow$ & R@1@0.7 $\uparrow$ & mIoU $\uparrow$ \\
        \midrule
        % TODO results pending for Gemini and GPT-5.6.
        % TODO confirm which Gemini version was evaluated and chronological placement of both models.
        % TODO add model references when available. No Gemini or GPT keys exist in the current bib files.
        % TODO describe closed model grounding prompts, video input sampling, and timestamp output parsing in Setup.
        % Gemini & -- & -- & -- & -- & -- & -- & -- \\
        % GPT-5.6 & -- & -- & -- & -- & -- & -- & -- \\
        % \midrule
        VideoChat-T~\citep{zeng2025timesuite} & 0.0 & 0.0 & 0.0 & 0.1 & 0.0 & 0.0 & 0.0 \\
        % ReVisionLLM~\citep{hannan2025revisionllm} & -- & -- & -- & -- & -- & -- & -- \\
        Time-R1~\citep{wang2025timer1} & 0.0 & 0.0 & 0.0 & 0.1 & 0.0 & 0.0 & 0.0 \\
        UniTime~\citep{li2025unitime} & 0.5 & 0.1 & 0.0 & 6.0 & 3.9 & 2.1 & 4.8 \\
        UniversalVTG~\citep{an2026universalvtg} & 5.8 & 2.7 & 1.2 & 22.1 & 14.6 & 8.7 & 16.3 \\
        TimeLens~\citep{zhang2026timelens} & 0.0 & 0.0 & 0.0 & 0.4 & 0.3 & 0.1 & 0.3 \\
        TimeLens2~\citep{zhu2026timelens2} & 1.1 & 0.8 & 0.3 & 10.0 & 6.9 & 3.9 & 7.4 \\
        Movie RAG~\citep{kim2026narrascene} & 7.1 & 4.1 & 2.0 & 25.8 & 19.1 & 12.2 & 18.5 \\
        % Single pass verification (27B) & -- & -- & -- & -- & -- & -- & -- \\
        \midrule
        Movie RAG + AWS (Ours, 4B) & 10.6 & 6.4 & 3.0 & 31.3 & 23.6 & 15.1 & 22.5 \\
        Movie RAG + AWS (Ours, 27B) & 15.0 & 8.8 & 3.7 & 37.0 & 28.2 & 17.9 & 26.8 \\
        % Movie RAG + reranker + AWS (Ours, 27B) & 17.1 & 9.6 & 4.1 & 39.1 & 29.3 & 18.4 & 28.2  \\
        \bottomrule
    \end{tabular}%
    }
    \vspace{-5pt}
\end{table}

\subsection{Main Results}

Public models that claim long-video temporal grounding struggle on \proposedBench{}, with UniversalVTG~\citep{an2026universalvtg} highest at 16.3 mIoU and several others near zero, indicating that full-movie confusable grounding is hard rather than that these models are weak in general.
Separating retrieval and localization is stronger, and Movie RAG reaches 18.5 mIoU, the best baseline here.
% TODO Confirm public model input protocols for a matched fps1 comparison.

Holding this baseline's retriever and localizer fixed and adding no task-specific training, Agentic Window Search raises mIoU to 26.8 with the 27B verifier and to 22.5 with the 4B verifier, and R@1@0.5 from 19.1 to 28.2.
AWS also improves pair success, raising PairSuccess@0.5 from 4.1\% to 8.8\%, so it more often grounds both confusable moments of a pair, though the low value shows that telling confusable moments apart stays hard.
It closes 24\% of the gap from Movie RAG to the oracle retriever reference of 53.8 (Table~\ref{tab:oracle_neighborhoods}, row~6).
These gains come entirely from checking candidate evidence, with the retriever, the localizer, and their weights unchanged, so the retrieval bottleneck identified in Section~\ref{sec:diagnosis} can be relieved without retraining any component. A single 27B pass over the top-five windows already lifts mIoU to 25.1, direct evidence checking working as the diagnosis predicts, and cascaded inspection raises it further to 26.8 (Section~\ref{sec:analysis}).

\subsection{Analysis of Agentic Window Search}
\label{sec:analysis}

% The Full row uses 27b-adjudicated-qwenfps1-mem.
% Dropped columns are Gold-chunk acc., Coverage, Inspect calls, Expansions, Frames, and Latency (s), in that order.
% Embedding top-1 dropped column values are 32.0 / 31.0 / -- / -- / -- / --. Its removed mIoU is 18.5, retained in Table 4.
% Full (27B with memory) dropped column values are 45.6 / 44.8 / -- / -- / -- / --.
% 27B w/o memory dropped column values are 44.9 / 44.3 / -- / -- / -- / --.
% 4B with memory dropped column values are -- / -- / -- / -- / -- / --.
% 4B w/o memory dropped column values are -- / -- / -- / -- / -- / --.
% M=0 dropped column values are -- / -- / -- / -- / -- / --.
% Select to top-1 dropped column values are -- / -- / -- / -- / -- / --.
\begin{wraptable}{r}{0.38\linewidth}
    \vspace{-38pt}
    \centering
    \small
    \caption{\textbf{Each AWS component helps, and reranking or fine-tuning the retriever adds further gains.} mIoU for a single change from the Full 27B run.}
    \vspace{-2mm}
    \label{tab:verification_ablation}
    \setlength{\tabcolsep}{2pt}
    \begin{tabular}{lr}
        \toprule
        Method & mIoU \\
        \midrule
        Full (27B) & 26.8 \\
        \midrule
        \multicolumn{2}{l}{\textit{Remove a component}} \\
        $\quad$w/o memory & 26.1 \\
        $\quad$w/o inspection & 25.1 \\
        \textcolor{gray}{$\quad$w/o both (Movie RAG)} & \textcolor{gray}{18.5} \\
        \midrule
        \multicolumn{2}{l}{\textit{Add a component}} \\
        $\quad$w/ Reranker & 28.2 \\
        $\quad$w/ Retriever FT & 30.6\\
        \bottomrule
    \end{tabular}
    \vspace{-2mm}
\end{wraptable}
Table~\ref{tab:verification_ablation} ablates the Full 27B configuration one change at a time, with the retriever and localizer fixed. Removing character memory lowers mIoU from 26.8 to 26.1, and removing the cascaded inspection of Section~\ref{sec:agentic_verification}, so that the verifier instead selects among the five candidate windows and their subtitles in a single inference, lowers it to 25.1, so both components contribute. A single inference already reads the candidate video but judges each window against the whole query, whereas decomposing the query and checking each sub-query against one window at a time keeps the distinguishing detail from being absorbed into the shared content, which accounts for the gain from 25.1 to 26.8. Removing both returns the pipeline to Movie RAG at 18.5 mIoU, so character memory and inspection together account for the gain. The retrieval fixes of Section~\ref{sec:diagnosis_retrieval_fixes} also help when placed before AWS, feeding the same verifier reranked or fine-tuned candidates while keeping the memory, localizer, and expansion budget fixed. Reranking reaches 28.2 mIoU (Table~\ref{tab:verification_ablation}), and fine-tuning the retriever reaches 30.6, each above the fix on its own in Table~\ref{tab:oracle_neighborhoods} and above AWS alone. Reranking reorders the candidates and fine-tuning rescores them, while AWS instead checks their evidence, so the retrieval fixes and AWS are complementary. Additional retriever and localizer combinations, MomentSeeker results, and qualitative examples are in Appendices~\ref{sec:app_additional} and~\ref{sec:app_qualitative}.

\vspace{-3pt}
\section{Conclusion}
\vspace{-3pt}
\label{sec:conclusion}

Grounding confusable movie moments requires checking the evidence that tells them apart rather than the content they share. We introduced \proposedBench{}, a benchmark that pairs confusable movie moments and grounds each query independently across the full movie. Diagnosing a retrieve-then-localize pipeline, we found window retrieval to be the main bottleneck, and that the retriever separates a moment from its confusable partner far less reliably than from an unrelated one, because its choice follows the content the two share rather than the detail that sets them apart. Together, these results suggest the principal failure is not insufficient retrieval capacity but insufficient use of the discriminative evidence that separates otherwise similar moments. Direct verification (Agentic Window Search) substantially reduces this failure even with the retriever and localizer fixed, showing the diagnosis is actionable. Gains remain bounded, as localization stays imperfect even with the best overlapping chunk, and the low PairSuccess shows grounding both queries in a pair is still hard. We hope \proposedBench{} guides future work that improves both window selection and boundary prediction and generalizes beyond a single retrieve-then-localize pipeline.

% The current evaluation leaves the generality and efficiency of this approach unresolved. It covers a single retriever and localizer pair and two verifiers from the Qwen family. Fixed chunks impose a coverage ceiling even with oracle selection, motivating window expansion without guaranteeing complete coverage under a finite budget. Agentic inspection adds inference cost, which must be weighed against grounding quality. Errors remain even when the gold chunk is supplied, so improved window selection does not eliminate the need for more accurate localization.

\section*{AI Use Statement}

Large language and multimodal models are components of the studied system. We use Gemini 3.1 Pro for benchmark query generation as described in Section~\ref{sec:benchmark} and Appendix~\ref{sec:app_benchmark}, Qwen3.6 27B for character memory construction in Appendix~\ref{sec:app_memory}, and Qwen verifiers for Agentic Window Search in Sections~\ref{sec:method} and~\ref{sec:experiments}. The verifier prompts are provided in Appendix~\ref{sec:app_prompts}.

\section*{Ethics Statement}

Movies and subtitles are sourced from ALLVB, and LLM generated queries are human verified by three of the authors. The benchmark concerns movie content and does not target private individuals. Character memory stores reference frames and names from the movies themselves. Our base pipeline uses pretrained models without fine tuning. We release queries, timestamps, and code in the supplementary material, but do not redistribute the videos.% Benchmark annotations, comprising queries and timestamps, are released under CC BY 4.0. Potential misuse concerns video search, including retrieving movie scenes outside their narrative context.

\section*{Reproducibility Statement}

Section~\ref{sec:benchmark} and Appendix~\ref{sec:app_benchmark} describe benchmark construction and human verification. The Setup and Evaluation paragraphs in Section~\ref{sec:experiments} specify models, sampling, thresholds, and metrics. Appendices~\ref{sec:app_public_inputs},~\ref{sec:app_memory}, and~\ref{sec:app_pipeline_combos} describe the input configurations of public baselines, character memory construction, and pipeline configurations. Appendix~\ref{sec:app_prompts} provides the verifier prompts. The supplementary material contains the Agentic Window Search code, an evaluation script, the \proposedBench{} annotations, and the training data of Section~\ref{sec:diagnosis_retrieval_fixes}.

\bibliography{references}
\bibliographystyle{iclr2027_conference}

\clearpage
\appendix
% Show only appendix sections and subsections in the contents.
\addtocontents{toc}{\protect\setcounter{tocdepth}{2}}
\begingroup
    \renewcommand{\contentsname}{Appendix Contents}
    \setcounter{tocdepth}{-1}
    \tableofcontents
\endgroup
\vspace{6pt}
\renewcommand{\thetable}{\Alph{table}}
\renewcommand{\thefigure}{\Alph{figure}}
\setcounter{table}{0}
\setcounter{figure}{0}

\section{Related Work}
\label{sec:related-work}

\paragraph{Temporal Grounding Benchmarks.}
Benchmarks cover movie grounding, localized television question answering, and long video comprehension~\citep{soldan2022mad,lei2018tvqa,lei2020tvqaplus,lei2020tvr,tapaswi2016movieqa,wang2025lvbench}, alongside audiovisual temporal understanding~\citep{geng2025longvale,chao2026jointavbench}.
Evaluation also tests binding within videos and grounding under compositional event conditions~\citep{saravanan2025velociti,zou2026comet}.
CoMET~\citep{zou2026comet} builds queries with an MLLM from sampled temporal and spatial attribute combinations, then grounds all matching events in long videos across multiple domains.
Some benchmark construction pipelines remove repeated events and separate search from localization in their analysis~\citep{seo2026extremewhenbench}.
To complement binding and event condition tests, our benchmark pairs confusable queries for full movie search, grounding each independently without the alternative query as input, so that success requires telling the paired moments apart.

\paragraph{Video Grounding and Agentic Search.}
Grounding methods develop hierarchical retrieval, compositional alignment, and negative ranking~\citep{zala2023hirest,li2022compositional,cheng2024shine}, as well as temporal representations and grounding supervision~\citep{guo2025trace,zeng2025timesuite,deng2025seq2time,wang2025timer1,zhang2026timelens,li2025unitime,zhu2026timelens2,an2026universalvtg}.
Search can narrow candidate windows through recursive localization, hierarchical alignment, or audiovisual guidance~\citep{hannan2025revisionllm,xu2026hitea,barrios2023guidance}.
Movie segmentation and scene representations organize retrieval units~\citep{huang2020movienet,rao2020lgss,chen2023movies2scenes}.
Movie RAG~\citep{kim2026narrascene} partitions movies into chunks, retrieves scenes by query similarity using the Qwen3-VL-Embedding~\citep{li2026qwen3} 2B model, and passes candidates to a temporal localizer.
We retain this structure and insert window search between retrieval and localization.
Agents gather evidence through tools and retrieval~\citep{wang2024videoagent,fan2024videoagent,ma2025drvideo,ataallah2024goldfish,shen2025vgent}, adapt observations to queries~\citep{wang2025videotree,yang2025vca,lin2026videoseek,zhang2026lens,xing2026omniagent}, and verify temporal candidates or decomposed cues~\citep{liu2026videomind,guo2025logicinframes}.
Complementary methods align subtitles and video for retrieval~\citep{lei2020tvr,li2020hero,zhang2021reloclnet,yoon2022squidnet}, link character references and speakers to visual evidence~\citep{lei2020tvqaplus,lee2025civlan,ouyang2026multispeaker}, and preserve identity or narrative context through memory and character prompting~\citep{song2024moviechat,han2023autoad2,xie2024autoadzero,ji2025idavlm,hahm2026storyteller}.
We combine dialogue-anchored character references with candidate inspection and bounded expansion to distinguish confusable moments without task-specific training.
The agent checks evidence and adjusts windows before passing its selection to a fixed localizer.

\section{Benchmark Construction and Verification}
\label{sec:app_benchmark}

\subsection{Query Pair Generation}
\label{sec:app_construction}

We source the 46 full-length movies from ALLVB~\citep{tan2025allvb}, whose official release provides movie files for research use.

We construct \proposedBench{} using Gemini 3.1 Pro~\citep{google2026gemini31pro} for query generation through three stages, namely
subtitle-anchored clip segmentation, event-level attribute generation, confusable query pair generation.

\begin{figure}[!t]
    \centering
    \includegraphics[width=\linewidth]{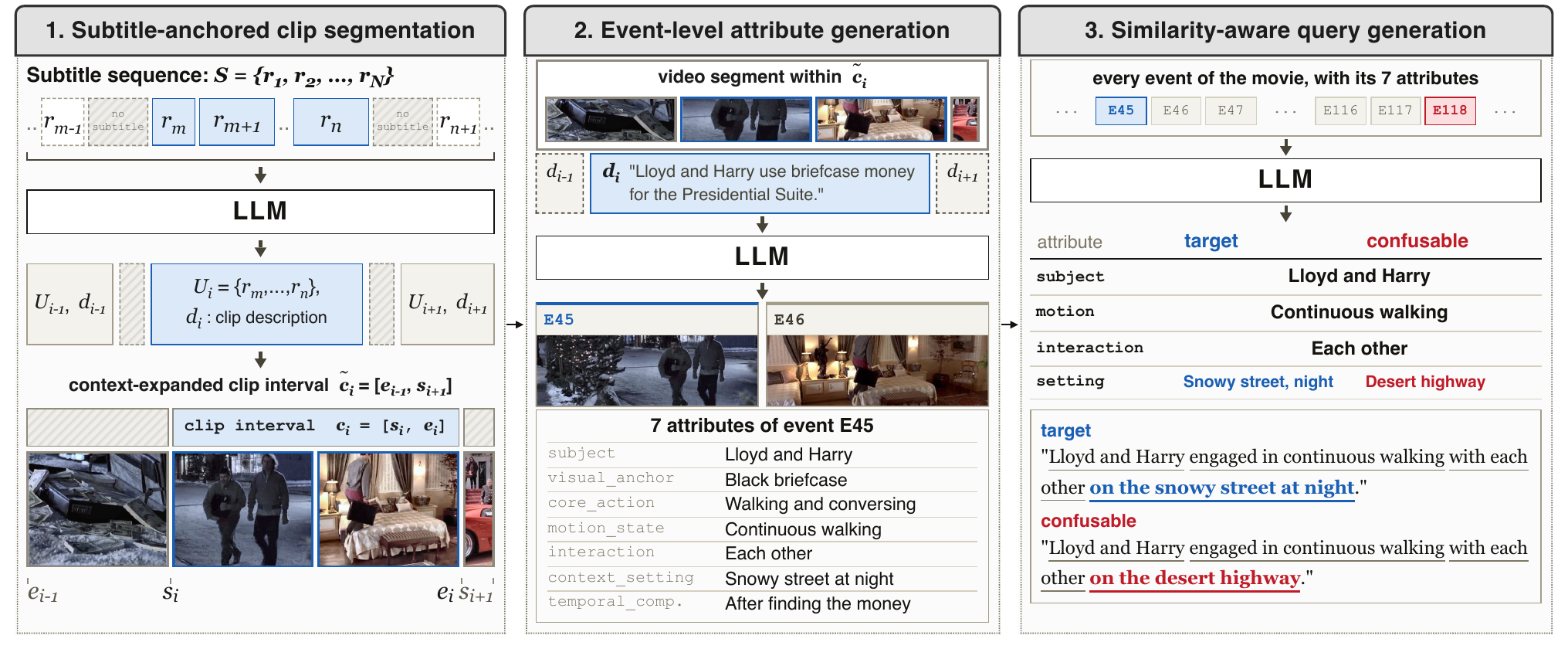}
    \caption{\textbf{Dataset construction pipeline.} Subtitle-anchored clip segmentation and description generation are followed by context expansion, event and attribute generation, and confusable query pair generation. The figure uses $R$ and $C_i$ for the subtitle sequence and cue groups, denoted $S$ and $U_i$ in the text.}
    \label{fig:data_construction}
    % TODO update figure notation to match text (subtitle sequence S, cue groups U_i)
\end{figure}

\paragraph{Subtitle-anchored clip segmentation.}

We first represent each movie subtitle as an ordered sequence of subtitle cues. A subtitle cue is the smallest temporal unit provided by the subtitle file and consists of an index, a timestamp interval, and its corresponding text. We denote the $k$-th subtitle cue as $r_k$. For example, a subtitle entry is represented as a subtitle cue $r_{169}$ containing its corresponding timestamp interval and text. Given an ordered subtitle sequence $S=\{r_1,r_2,\dots,r_N\}$, we employ an LLM-based scene segmentation process to group consecutive subtitle cues into semantically coherent clips. The LLM analyzes the subtitle content and temporal continuity of adjacent cues, merging cues that belong to the same scene, conversation, or continuous narrative episode into a single clip. Specifically, the $i$-th subtitle-derived clip is defined as a contiguous group of cues $U_i=\{r_m,r_{m+1},\dots,r_n\}$, where $m$ and $n$ denote the first and last cue indices assigned to the clip. The temporal interval of the clip is determined by the start time of the first cue and the end time of the last cue, i.e., $c_i=[t_m^{s},t_n^{e}]$.
Along with the clip boundaries, the LLM generates a concise natural-language description \(d_i\) for each clip, summarizing the main action or narrative content expressed by its grouped subtitle cues. Thus, each subtitle-derived clip \(c_i\) is paired with a corresponding clip description \(d_i\), which is subsequently used as semantic context during event-level processing.
After obtaining subtitle-derived clips, we construct a context-expanded clip video for each clip. Given an original clip $c_i=[s_i,e_i]$, we define its expanded interval as $\tilde{c}_i=[e_{i-1},s_{i+1}]$, where the preceding and following clip boundaries provide additional visual context. For the first and last clips, the movie boundaries are used when a preceding or following clip is unavailable. The original clip interval $c_i$ is preserved inside the expanded interval $\tilde{c}_i$, together with its relative temporal position. Therefore, the expanded video contains the subtitle-free visual context immediately before and after the original clip without absorbing the neighboring clips themselves.

\paragraph{Event-level attribute generation.}

For each clip \(c_i\), the multimodal model is provided with the context-expanded clip video \(\tilde{c}_i\), the location of the original clip interval \(c_i\) within it, and the subtitle-derived descriptions of the preceding, current, and following clips, \(d_{i-1}\), \(d_i\), and \(d_{i+1}\), respectively. The clip descriptions provide high-level narrative context for interpreting the visual content and its relation to neighboring clips, while the event boundaries and attributes are grounded in the video.
% The preceding clip description serves two roles in handling the subtitle-free gap between adjacent clips. It helps determine whether a meaningful event in the gap belongs to the preceding clip, preventing duplicate event generation, and, even when no distinct event exists in the gap, helps localize the onset of the current event beyond the subtitle boundary.
Events are primarily generated from the visual content within the original clip interval. If the original clip contains multiple visually separable events, it is decomposed into event-level temporal segments. Since subtitle-derived clip boundaries may not exactly align with actual visual event boundaries, the surrounding visual context is used to refine the temporal extent of events and to identify additional visually meaningful events near clip boundaries. For the subtitle-free gap preceding the original clip, the preceding clip description \(d_{i-1}\) helps determine whether the observed content belongs to the preceding clip. Content associated with the preceding clip is excluded to avoid duplication, whereas a visually meaningful event that is neither associated with the preceding clip nor part of the current clip event is generated as an additional event for the current clip. In contrast, the visual context after the original clip interval is utilized as additional information for understanding the original clip, while newly occurring visually separable events within this context are not added to the current clip. These events are assigned to the corresponding following clip to prevent duplicate event generation across adjacent context-expanded clip videos.

For each event, we generate attributes belonging to three general semantic families.
Visual information represents object-level and scene-level visual cues associated with the event.
Action information describes the observable behaviors and motions within the event.
Temporal-relation information captures the dependencies and temporal relationships between the event and its surrounding contexts. In implementation, these semantic families are represented using seven structured slots, namely subject, visual anchor, core action, motion state, interaction cause, context setting, and temporal composition.
Specifically, subject, visual anchor, and context setting constitute Visual information, while core action and motion state constitute Action information.
The interaction/cause and temporal composition slots represent Temporal-relation information, capturing the cause of the event and its temporal relationship with nearby events. The resulting representation provides a structured basis for comparing events across the entire movie.

\paragraph{Confusable query pair generation.}

We next identify confusable event pairs based on the event-level attributes. We select event attributes from temporally distant clips to construct confusable event pairs. For every retained pair, we select exactly three shared attributes and one discriminative attribute. Shared attributes must be semantically compatible for both events and appear independently in each event. The discriminative attribute captures the principal difference needed to distinguish the target from its confusable counterpart. We then generate two parallel descriptive queries using the same four attribute roles. The queries are written as direct event descriptions rather than retrieval instructions or questions. This parallel construction ensures that the principal linguistic difference between the two queries reflects the intended event-level contrast rather than incidental variation in sentence structure. Finally, each pair is converted into temporal-grounding instances. Each instance consists of a target query $q_t$, a paired confusable query $q_c$, and their corresponding temporal intervals $\tau_t$ and $\tau_c$.

\subsection{Human Verification}
\label{sec:app_verification}

\begin{figure}[t]
    \centering
    \begin{subfigure}[t]{0.49\linewidth}
        \centering
        \includegraphics[width=\linewidth]{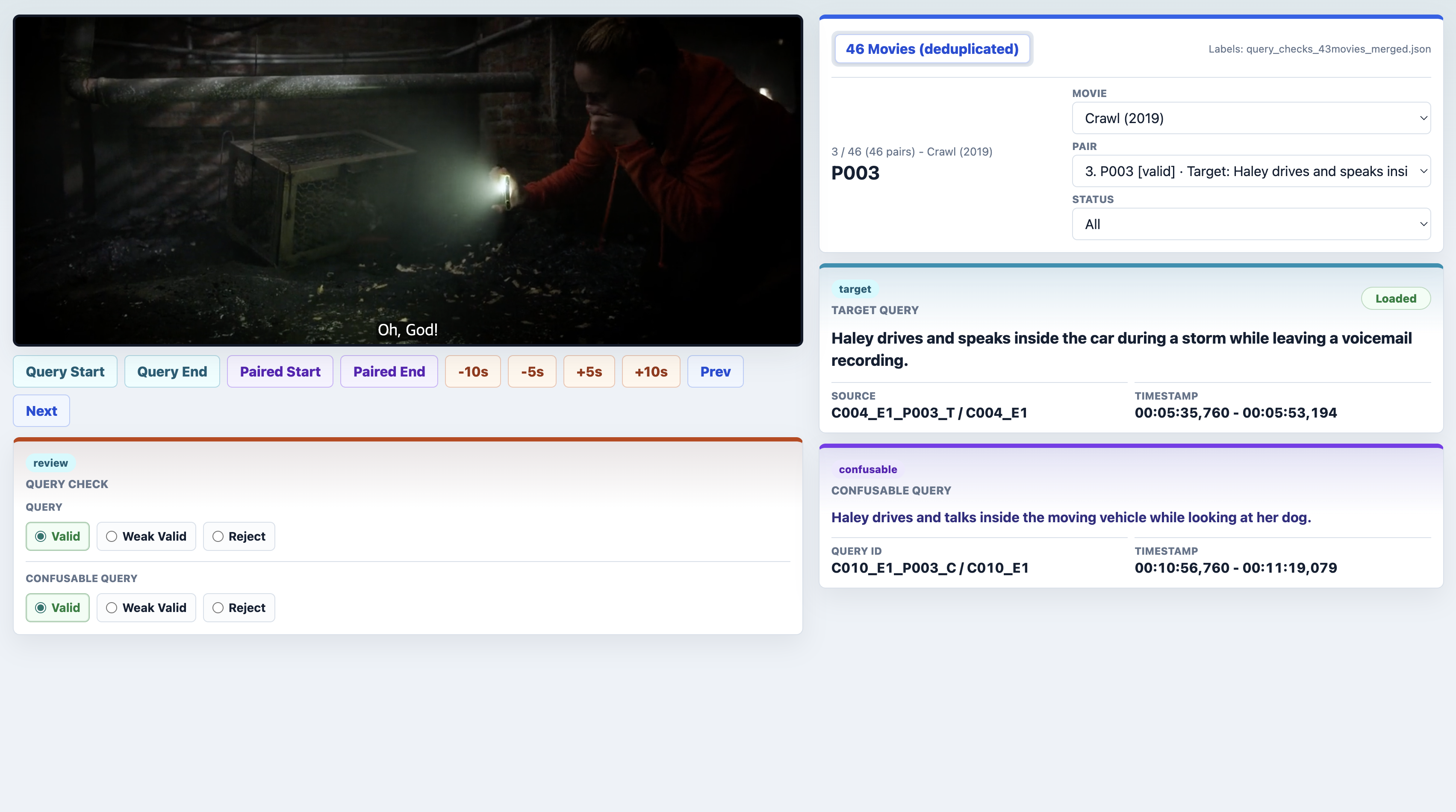}
        \caption{Initial verification and second-pass review.}
        \label{fig:annotation_ui_check}
    \end{subfigure}
    \hfill
    \begin{subfigure}[t]{0.49\linewidth}
        \centering
        \includegraphics[width=\linewidth]{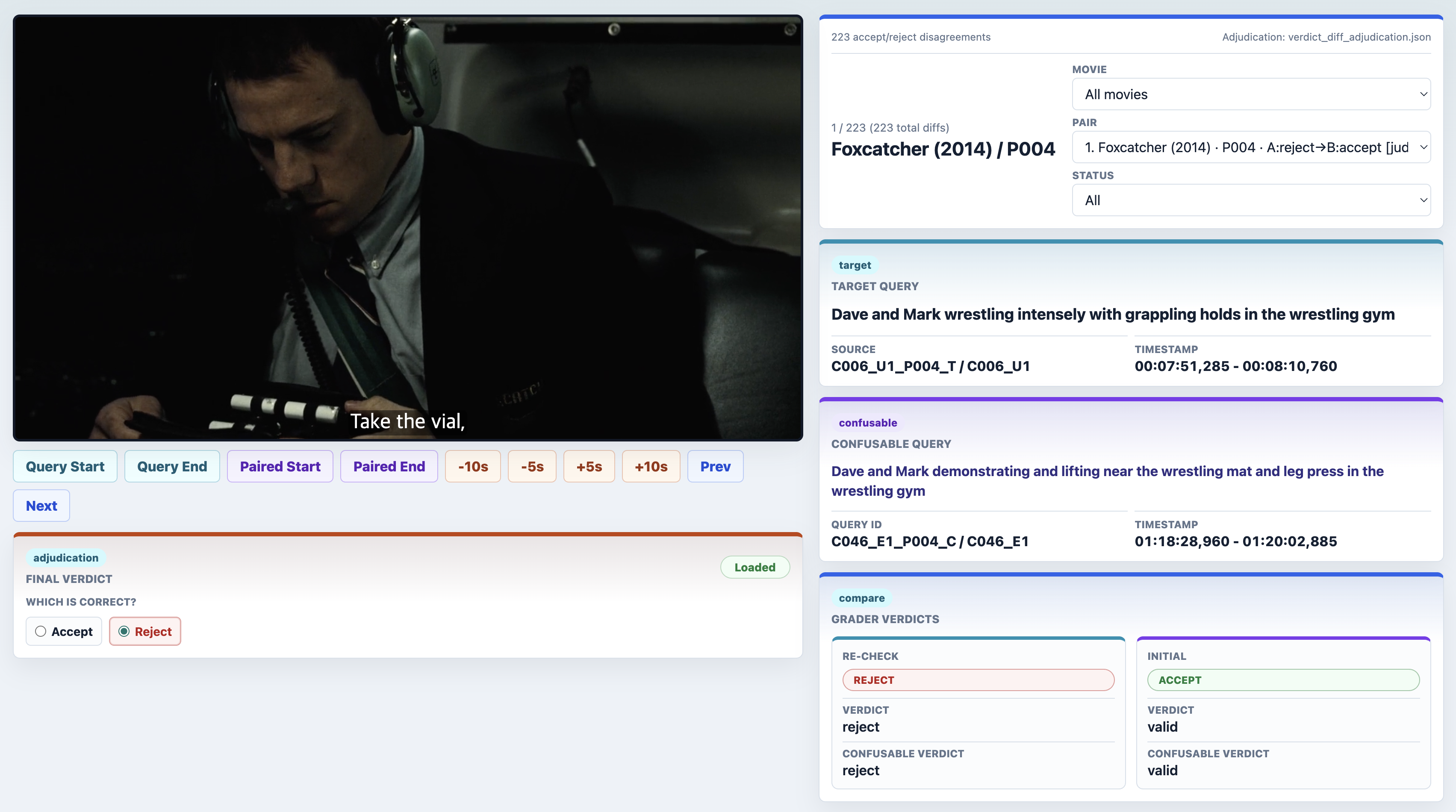}
        \caption{Adjudication of disagreements.}
        \label{fig:annotation_ui_adjudication}
    \end{subfigure}
    \caption{\textbf{Verification interface.} (a) The target query and its confusable counterpart each receive a separate verdict, both in the initial verification and in the later review by another annotator. (b) For pairs whose acceptance outcomes disagree, the two preceding verdicts are displayed together for the resolving decision.}
    \label{fig:annotation_ui}
\end{figure}

Reviewers used a web interface that plays the movie with subtitles, shows both queries with their proposed intervals and structured attributes, and allows jumping to interval boundaries before recording a decision (Figure~\ref{fig:annotation_ui}).

\begin{figure}[t]
    \centering
    \setlength{\fboxsep}{10pt}
    \fcolorbox[gray]{0.5}{0.95}{%
    \begin{minipage}{\dimexpr\linewidth-2\fboxsep-2\fboxrule\relax}
        \textbf{Annotator guideline}
        \par\medskip
        Accept an annotation only if all of the following are satisfied.
        \begin{enumerate}
            \item \textbf{Query validity.} The event or content described by the query clearly occurs in the video and is unambiguous.
            \item \textbf{Temporal relevance.} The content described by the query is clearly visible or can be confidently inferred within the annotated GT interval.
            \item \textbf{Temporal accuracy.} The GT interval tightly covers the relevant event without substantial irrelevant content or missing important portions.
            \item \textbf{Uniqueness.} There is no other substantially different temporal interval that is equally valid for the same query.
            \item \textbf{Evaluation validity.} A temporal grounding prediction that exactly matches the GT interval ($\mathrm{IoU}=1$) should clearly represent a more accurate localization of the query than predictions outside or substantially shifted from the GT interval.
        \end{enumerate}
        Reject the annotation if any of these conditions are not clearly satisfied.
        When uncertain or when multiple intervals are equally plausible, choose \texttt{reject}.
    \end{minipage}%
    }
    \caption{\textbf{Annotator guideline.} Acceptance criteria shown to reviewers during human verification.}
    \label{fig:annotation_guideline}
\end{figure}

The acceptance criteria are shown in Figure~\ref{fig:annotation_guideline}.
Reviewers judge the two members of each pair independently, and a pair is retained only when both queries are accepted, otherwise the whole pair is removed. Reviewers also confirm that the two queries form a meaningful confusable pair, sharing the intended attributes while remaining distinguishable by the discriminative one. Of the 4,080 queries automatically generated across the 46 movies, 3,014 were retained after verification.
Three annotators divided the 46 movies among themselves for the initial verification.
Each annotator then reviewed six movies assigned to another annotator without access to the initial decisions, with B reviewing A's movies, C reviewing B's movies, and A reviewing C's movies, covering 18 movies in total.
For binary agreement, \texttt{valid} is treated as acceptance.
Inter-annotator agreement on pair acceptance before adjudication, measured by Cohen's $\kappa$, was 0.771, indicating substantial agreement on pair acceptance rather than independently assigned temporal boundaries.
Disagreements were resolved by the third annotator who had not provided either preceding judgment.
Annotators judged each query against the acceptance criteria (Figure~\ref{fig:annotation_guideline}) rather than assigning new boundaries.
Table~\ref{tab:iaa} reports agreement before adjudication.

\begin{table}[t]
    \centering
    \caption{\textbf{Reviewers agree on pair acceptance before adjudication.} Agreement before adjudication by annotator assignment, where arrows mark the initial reviewer followed by the reviewer performing the blind check.}
    \label{tab:iaa}
    \small
    \setlength{\tabcolsep}{4pt}
    \begin{tabular}{lccccc}
        \toprule
        \makecell{Reviewer\\assignment} & \makecell{Movies} & \makecell{Compared\\pairs} & \makecell{Agreement\\(\%)} & $\kappa$ & \makecell{Disagreements} \\
        \midrule
        A $\to$ B & 6 & 282 & 91.1 & 0.775 & 25 \\
        B $\to$ C & 6 & 269 & 90.0 & 0.778 & 27 \\
        C $\to$ A & 6 & 259 & 90.3 & 0.757 & 25 \\
        Overall & 18 & 810 & 90.5 & 0.771 & 77 \\
        \bottomrule
    \end{tabular}
\end{table}

\section{Baseline and Retriever Details}
\label{sec:app_baseline}

\subsection{Input Configurations of Public Grounding Models}
\label{sec:app_public_inputs}
Every baseline receives the complete movie. The released code of Time-R1
and TimeLens-8B cap a video at 768 frames, and that of TimeLens2-8B at 2,048 frames, which
would sample a median-length movie only once every 9 or 3.4 seconds. We therefore raise the
frame limit of these three models to 12,000 frames, which no movie in \proposedBench{}
reaches, so that each of them samples the full movie at 1\,fps. VideoChat-T cannot be
extended in this way, because its released checkpoint fixes the length of its visual input;
it therefore samples uniformly over the full movie within that fixed length. UniTime extracts
features from 2,048 uniformly sampled frames and runs its released coarse-to-fine procedure,
selecting a temporal window from the full movie before localizing within it. UniversalVTG
operates on cached full-movie features extracted at 1\,fps.

\subsection{Retriever Analysis}
\label{sec:app_retriever}

We break down confusable gold selection by contrast attribute for Qwen3-VL-Embedding and WeMM-Embedding~\citep{wemm-embedding} on the same full set of 3,014 queries, sampled at one frame per second under the Random and Confusable definitions of Section~\ref{sec:diagnosis_bottleneck}, in Table~\ref{tab:app_twoway_attr}.

\begin{table}[t]
    \centering
    \small
    \caption{\textbf{Confusable gold selection by contrast attribute.} Accuracy is reported in percent. Both retrievers use the same full 3,014 query set sampled at one frame per second. Each $N$ entry lists the Qwen3-VL-Embedding and WeMM-Embedding query counts in that order, separated by a slash. All attribute groups are available for both retrievers.}
    \label{tab:app_twoway_attr}
    \begin{tabular}{lccc}
        \toprule
        Contrast attribute & $N$ & Qwen3-VL-Embedding & WeMM-Embedding \\
        \midrule
        setting & 1,310 / 1,310 & 77.3 & 77.8 \\
        subject & 382 / 382 & 66.8 & 67.5 \\
        core action & 566 / 566 & 66.1 & 65.9 \\
        motion & 212 / 212 & 56.6 & 61.3 \\
        interaction & 196 / 196 & 67.9 & 69.4 \\
        visual anchor & 336 / 336 & 69.6 & 70.2 \\
        temporal composition & 12 / 12 & 66.7 & 58.3 \\
        \bottomrule
    \end{tabular}
\end{table}

Attribute groups differ in sample size and movie composition, so these descriptive results do not isolate the effects of individual attributes.

\begin{figure}[ht]
    \centering
    \includegraphics[width=0.55\linewidth]{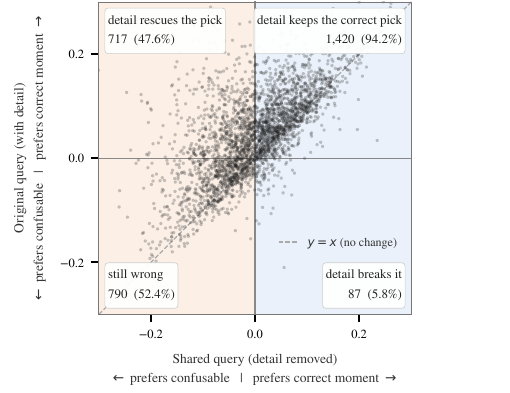}
    \caption{\textbf{Shared content decides most retrieval picks.} Each point is one query, placed by its margin under the shared query ($x$) and the original query ($y$). Points stay near $y=x$, so the detail rarely changes the pick.}
    \label{fig:app_shared_scatter}
\end{figure}

Figure~\ref{fig:app_shared_scatter} extends the shared-query analysis of Section~\ref{sec:diagnosis_shared_query}, which Figure~\ref{fig:shared_query} illustrates with one pair, by showing the margin of every query under the shared and original queries.

\section{Agentic Window Search Details}
\label{sec:app_aws}

\subsection{Cascaded Inspection}
\label{sec:app_cascaded}

Algorithm~\ref{alg:verifier} states the full cascaded inspection procedure that Section~\ref{sec:agentic_verification} describes, inspecting ranked candidates in order and expanding a window within a fixed budget until one is accepted or a fallback selection is made.

\begin{algorithm}[t]
    \caption{Cascaded Inspection with Bounded Expansion for one query.
    \textsc{Inspect} checks each sub-query in $D$ using fixed references $R$.
    \textsc{Accept} returns $W$, \textsc{Reject} advances, and
    \textsc{Expand} reveals adjacent evidence within budget.}
    \label{alg:verifier}
    \small
    \begin{algorithmic}[1]
        \REQUIRE Query $q$, ranked $C_{1:K}$ ($K\geq1$), movie, character memory
        \REQUIRE At most $M$ expansions per candidate, each at most $\Delta_{\max}$
        \ENSURE A window $W$ for the downstream localizer
        \STATE $P\gets$ \textsc{MatchCharacters}$(q)$ \COMMENT{name lookup in memory}
        \STATE $(D,R)\gets$ \textsc{Decompose}$(q,P)$ \COMMENT{single model call}
        \STATE $H\gets[\,]$ \COMMENT{rejected windows and histories}
        \FOR[stop at first acceptance]{each $C_r$ in retrieval order}
            \STATE $W\gets C_r.\mathrm{window}$
            \STATE Initialize $h$ with $q,D,R$ and observations from $C_r$
            \FOR[verify with bounded expansion]{$n=0,\ldots,M$}
                \STATE $(o,h)\gets$ \textsc{Inspect}$(h,\mathrm{allow\_expand}=(n<M))$
                \IF{$n=M$ and $o.\mathrm{decision}=\textsc{Expand}$}
                    \STATE $o.\mathrm{decision}\gets\textsc{Reject}$
                \ENDIF
                \IF{$o.\mathrm{decision}=\textsc{Accept}$}
                    \STATE \textbf{return} $W$
                \ELSIF{$o.\mathrm{decision}=\textsc{Reject}$}
                    \STATE Append $(W,h)$ to $H$
                    \STATE \textbf{break}
                \ELSE
                    \STATE $(W,h)\gets$ \textsc{ExpandWindow}$(W,h,o,\Delta_{\max})$ \COMMENT{new band only}
                \ENDIF
            \ENDFOR
        \ENDFOR
        \STATE $W\gets$ \textsc{Select}$(q,D,H)$ \COMMENT{VLM selects among rejected}
        \STATE \textbf{return} $W$
    \end{algorithmic}
\end{algorithm}

\subsection{Character Memory Construction}
\label{sec:app_memory}

We construct character memory separately for each movie from its video, subtitles, and identifier without retrieval queries or ground truth timestamps.
Qwen3.6 27B~\citep{qwen36_27b}, used throughout construction, extracts names such as ``Alex Smith'' and ``Alex'' from subtitle blocks overlapping by eight entries.
Duplicate extractions sharing a normalized proposed name are combined, counting each supporting subtitle entry once.
A text only consolidation call uses these counts and sampled contexts to select recurring characters and propose groups of names that refer to the same person.
Each group's longest name serves as its provisional display name, while visual evidence is collected separately for every name.
Candidate name pairs come from these groups, unique matches between short and full names, or another text only call suggesting equivalence from dialogue.
For each name, matching subtitle entries are candidates for video inspection.
We shortlist up to 40 entries, combining lexical heuristic scores with temporal coverage.
Using surrounding dialogue, the model excludes mentions classified as references to third parties or possessive references and prioritizes statements identifying the speaker, introductions, and direct address suggesting participation.
We choose up to ten entries per name, at least $45$\,s apart, adding eligible weaker mentions if fewer than two strong candidates qualify.
Their timestamps determine which video intervals to inspect.

Each selected timestamp centers a $30$\,s reference clip with overlapping subtitles, uniformly sampled at $1$\,fps.
One independent call per reference clip links the name to a visible person and selects at most two frames.
When at least two reference clips yield candidates, one call jointly compares all selected frames using stable physical cues and identifies reference clips consistent with one character.
References must pass confidence checks for the name and identity and quality gates for each frame.
Frame identity confidence is the minimum of grounding scores and consistency scores across reference clips.
Pairwise comparison requires accepted references for both names.
Comparison decisions distinguish the same person in the same or different visual forms from different people, or indicate uncertainty.

\begin{figure}[t]
    \centering
    \includegraphics[width=\linewidth]{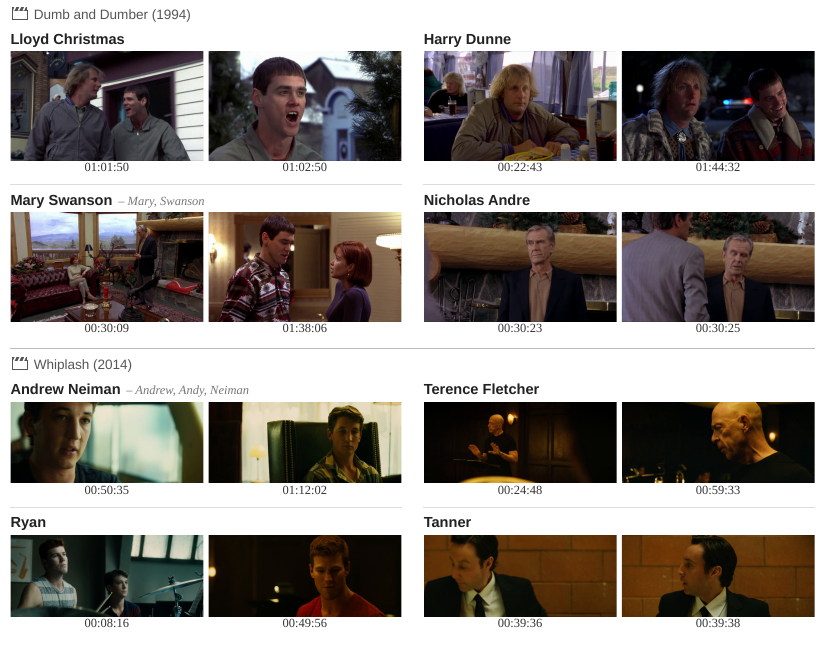}
    \caption{\textbf{Stored character memory examples.} Entries from two movies with two reference frames each and their timestamps. Display names use credited character names and the smaller italic names reproduce the other stored name records.}
    \label{fig:app_memory_examples}
\end{figure}

Identity resolution uses two stages of constrained union find.
With a confidence threshold of $0.80$, comparison decisions identifying different people first establish constraints that prohibit merging their names.
The first stage merges name groups supported by comparison decisions identifying the same person unless a prohibitive constraint connects their members.
Within each character group, the second stage starts with each name separate.
Accepted comparison decisions identifying different visual forms prohibit variant merges, while accepted comparison decisions identifying the same form join groups only when these constraints are respected.
Remaining groups define variants.
Conflicting merges of the same form are skipped, and results depend on processing order, which is not fixed.
Each variant pools accepted references and is retained only with support from at least two reference clips, storing up to four references in $\mathcal{M}$.
Figure~\ref{fig:app_memory_examples} shows character memory examples from two movies.

\section{Additional Experiments}
\label{sec:app_additional}

This appendix reports experiments beyond the main setup of Section~\ref{sec:experiments}, first varying the retriever and localizer and then testing transfer to another dataset.

\subsection{Robustness to the Retriever and Localizer}
\label{sec:app_pipeline_combos}

The main results fix Qwen3-VL-Embedding and TimeLens2. Table~\ref{tab:app_pipeline_combos} instead varies these two components, and AWS keeps a comparable gain when the retriever is changed to WeMM-Embedding.

\begin{table}[t]
    \centering
    \small
    \caption{\textbf{Pipeline configuration comparisons.} Baseline and Ours use the same retriever and localizer, and Gain denotes the change in mIoU. Ours uses the 27B verifier with character memory and the corresponding retrieval cache.}
    \label{tab:app_pipeline_combos}
    \setlength{\tabcolsep}{3pt}
    \begin{tabular}{llccccc}
        \toprule
        Retriever & Localizer & \shortstack{Baseline\\mIoU} & \shortstack{Ours\\mIoU} & Gain & \shortstack{Baseline\\R@1@0.5} & \shortstack{Ours\\R@1@0.5} \\
        \midrule
        Qwen3-VL-Embedding & TimeLens2 & 18.5 & 26.8 & +8.3 & 19.1 & 28.2 \\
        WeMM-Embedding & TimeLens2 & 17.9 & 26.8 & +8.9 & 18.8 & 28.3 \\
        Qwen3-VL-Embedding & UniversalVTG & 15.8 & 23.4 & +7.6 & 15.3 & 22.7 \\
        \bottomrule
    \end{tabular}
\end{table}

% The Qwen reference is \texttt{27b-adjudicated-qwenfps1-mem}, and the WeMM reference is \texttt{27b-adjudicated-wemm-mem}.
Both AWS configurations use the same 3,014 queries and their respective retrieval rankings.
% Table~\ref{tab:app_pair_thresholds} preserves the complete reported thresholds by run.
% Missing R@1@0.7 values are left pending, including the separate 4B memory report whose prediction identity is not yet linked.
% For WeMM, memory changes 27B PairSuccess@0.5 from 9.1\% to 8.7\%, and 4B PairSuccess@0.5 from 7.0\% to 6.1\%.
% These small differences do not establish a causal benefit or harm from memory, and a retriever change on the same benchmark is not external generalization.
A retriever change on the same benchmark is not external generalization.

% \begin{table}[t]
%     \centering
%     \small
%     \caption{\textbf{Reported fps1 query and pair results.} This table expands the comparison in Table~\ref{tab:frozen_public_models} with memory ablations and WeMM configurations. R denotes R@1 and P denotes PairSuccess. Within each retriever block, every row corresponds to a single reported run. Missing values remain unconfirmed.}
%     \label{tab:app_pair_thresholds}
%     \setlength{\tabcolsep}{3pt}
%     \begin{tabular}{lrrrrrrr}
%         \toprule
%         Method & mIoU & R@0.3 & R@0.5 & R@0.7 & P@0.3 $\uparrow$ & P@0.5 $\uparrow$ & P@0.7 $\uparrow$ \\
%         \midrule
%         \multicolumn{8}{l}{Qwen fps1} \\
%         Movie RAG & 18.5 & 25.8 & 19.1 & -- & 7.1 & 4.1 & 2.0 \\
%         Ours (4B, w/o memory) & 22.3 & 30.8 & 23.7 & -- & 9.6 & 5.8 & 2.5 \\
%         Ours (4B) & 22.5 & 31.3 & 23.6 & -- & 10.6 & 6.4 & 3.0 \\
%         Ours (27B, w/o memory) & 26.3 & 36.6 & 27.5 & -- & 14.7 & 8.8 & 3.6 \\
%         Ours (27B) & 26.8 & 37.1 & 28.2 & -- & 15.0 & 8.8 & 3.7 \\
%         \midrule
%         \multicolumn{8}{l}{WeMM fps1} \\
%         Movie RAG & 17.9 & 25.2 & 18.8 & -- & 7.0 & 4.2 & 1.4 \\
%         Ours (4B, w/o memory) & 22.3 & 31.2 & 23.5 & -- & 11.3 & 7.0 & 2.4 \\
%         Ours (4B) & 22.2 & 31.0 & 23.1 & -- & 10.6 & 6.1 & 2.4 \\
%         Ours (27B, w/o memory) & 26.5 & 37.0 & 28.0 & -- & 15.1 & 9.1 & 3.5 \\
%         Ours (27B) & 26.8 & 37.5 & 28.3 & -- & 15.3 & 8.7 & 3.3 \\
%         \bottomrule
%     \end{tabular}
% \end{table}

The fine tuned retriever, reranking, and their combination with AWS are covered in Sections~\ref{sec:diagnosis_retrieval_fixes} and~\ref{sec:analysis}.
Table~\ref{tab:app_larger_retriever} additionally reports the base pipeline with a larger Qwen3-VL-Embedding 8B retriever in place of the 2B used by Movie RAG, both with TimeLens2 and without AWS, under the same evaluation protocol as the main results. The public grounding models appear in Table~\ref{tab:frozen_public_models}.
% InfoNCE positive and negative units, trainable parameters, optimization, and checkpoint selection remain to be documented.

\begin{table}[t]
    \centering
    \small
    \caption{\textbf{A larger retriever gives only a small gain.} Replacing the 2B retriever of Movie RAG with a larger Qwen3-VL-Embedding 8B, both with TimeLens2 and without AWS, raises mIoU only from 18.5 to 20.1, so retriever scale alone does not close the gap. R denotes R@1 and P denotes PairSuccess.}
    \label{tab:app_larger_retriever}
    \setlength{\tabcolsep}{3pt}
    \begin{tabular}{lrrrrr}
        \toprule
        Retriever (with TimeLens2) & mIoU & R@0.3 & R@0.5 & R@0.7 & P@0.5 \\
        \midrule
        Qwen3-VL-Embedding 2B (Movie RAG) & 18.5 & 25.8 & 19.1 & 12.2 & 4.1 \\
        Qwen3-VL-Embedding 8B & 20.1 & 28.4 & 20.9 & 13.0 & 4.9 \\
        \bottomrule
    \end{tabular}
\end{table}

\subsection{Generalization to MomentSeeker}
\label{sec:app_momentseeker}

We evaluate whether AWS transfers to text only queries on MomentSeeker~\citep{yuan2025momentseeker} without task specific training.
The Movie RAG style baseline and Ours use the same retriever and TimeLens2 localizer without character memory.
We use our pipeline setting of 180\,s chunks, which differs from MomentSeeker's 10\,s chunk protocol.
We report the same grounding metrics as the main results, R@1 at 0.3, 0.5, and 0.7 and mIoU, counting a prediction as correct when it matches any ground truth interval at the specified IoU threshold following MomentSeeker. PairSuccess does not apply because MomentSeeker has no confusable pairs.
AWS transfers to this benchmark, raising mIoU from 19.9 to 24.8 with the 27B verifier (Table~\ref{tab:app_momentseeker}).

\begin{table}[t]
    \centering
    \small
    \caption{\textbf{Generalization to MomentSeeker~\citep{yuan2025momentseeker}.} Evaluation uses text only queries. Movie RAG and Ours use 180\,s chunks with the retriever and TimeLens2 fixed and no character memory, which differs from MomentSeeker's 10\,s chunk protocol. Grounding metrics follow the main results, and PairSuccess is omitted because MomentSeeker has no confusable pairs.}
    \label{tab:app_momentseeker}
    \setlength{\tabcolsep}{4pt}
    \begin{tabular}{lcccc}
        \toprule
        Method & mIoU $\uparrow$ & R@1@0.3 $\uparrow$ & R@1@0.5 $\uparrow$ & R@1@0.7 $\uparrow$ \\
        \midrule
        Movie RAG & 19.9 & 29.6 & 15.7 & 6.9 \\
        Movie RAG + AWS (4B) & 23.4 & 34.8 & 19.5 & 8.3 \\
        Movie RAG + AWS (27B) & 24.8 & 37.2 & 20.8 & 8.8 \\
        \bottomrule
    \end{tabular}
\end{table}

% \section{Inference Cost}
% \label{sec:app_cost}
%
% Table~\ref{tab:app_cost} retains placeholders for inference cost per query for Movie RAG and both verifier configurations. Costs require separate accounting for offline indexing and memory construction, online calls, unique and repeated frames, tokens, latency, hardware, precision, batching, caching, and retries. Task specific training is unnecessary for base AWS, but this does not establish a compute advantage.
%
% \begin{table}[t]
%     \centering
%     \small
%     \caption{\textbf{Inference cost.} Total calls count Decompose, Inspect, and Select, and offline memory construction is excluded.}
%     \label{tab:app_cost}
%     \begin{tabular}{lccc}
%         \toprule
%         Method & \shortstack{Inspect calls\\per query} & \shortstack{Total calls\\per query} & \shortstack{Seconds\\per query} \\
%         \midrule
%         Movie RAG & -- & -- & -- \\
%         Ours (4B) & -- & -- & -- \\
%         Ours (27B) & -- & -- & -- \\
%         \bottomrule
%     \end{tabular}
% \end{table}

\section{Additional Qualitative Examples}
\label{sec:app_qualitative}

\begin{figure}[t]
    \centering
    \includegraphics[width=\linewidth]{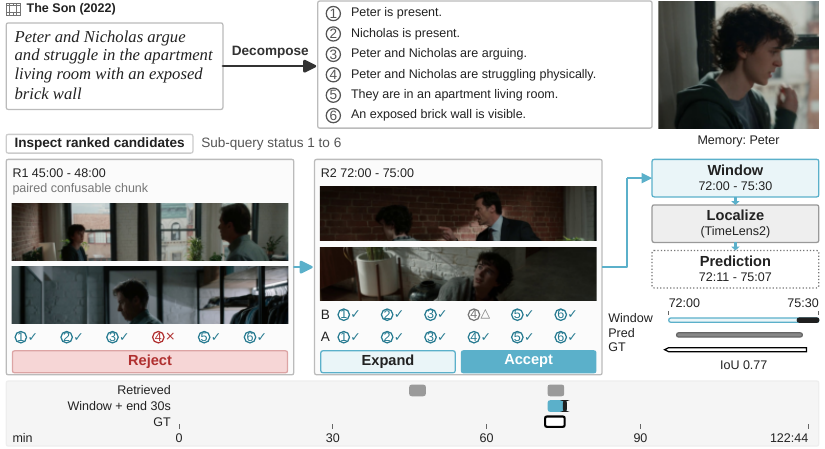}
    \caption{\textbf{Qualitative example with bounded expansion.} The paired confusable chunk is rejected because the physical struggle sub-query is false, while a 30\,s rightward expansion of the next candidate changes that sub-query from cutoff to true and leads to acceptance. Numbered chips show the recorded model status for each sub-query. Check marks, crosses, and triangles denote sub-queries judged true, false, or cut off at the window boundary, and B and A rows show statuses before and after expansion.}
    \label{fig:qualitative_theson}
\end{figure}

Figure~\ref{fig:qualitative_theson} illustrates a model-generated status change after bounded expansion in The Son. It explains the procedure without establishing a quantitative expansion effect.

\section{Prompts}
\label{sec:app_prompts}

These system prompts and user templates are used by the zero shot verifier for Decompose, Inspect, and Select. Inspect is named Look in the implementation, and EMIT corresponds to acceptance. Braced fields in user templates denote runtime substitutions, while JSON braces in system prompts are literal. Image content is attached separately at the positions described below. The serialized sub-queries exclude the internal character reference image field. Optional subtitle blocks are empty when no subtitles are available. Inspect retains prior model replies and expansion observations in its conversation history.
% Source files are decompose.py, look.py, and select.py in zero-shot_temporal_verifier/prompts.
% Source punctuation is normalized to ASCII inside verbatim blocks.

Decompose splits the query and selects character references using \texttt{query}, \texttt{cid}, \texttt{display\_name}, and \texttt{j}, repeating candidate names and thumbnail labels with each image following its label.

\begin{promptbox}{Decompose}
System prompt
You split a scene description into atomic statements.

Each statement will be checked against a stretch of video by asking: "is this true at any moment in here?" So each statement must describe something observable at a moment -- a person, an action, a place that is on screen, a line that is spoken.

Rules:
- Split anything that joins two claims, one claim per statement. Examples: "A and B are talking in the kitchen" becomes "A appears" / "B is visible" / "Two people are talking in a kitchen". "A runs down a dark hallway" becomes "A is running" / "A is in a dark hallway". Vary the wording; do not reuse one phrasing for every statement.
- Write about what is on screen at a moment, NOT about "the scene" as a whole. Write "Two people are in an underground lab", not "The scene takes place in an underground lab". Never begin a statement with "The scene", "The setting", "The segment", or "The video".
- Keep proper nouns exactly as written: if the description names a person or place, use that name in the statement rather than a generic description (write "Kaffee is shouting", not "A man is shouting").
- State only what the description states. Do not add conditions.
- Each atomic statement must be entailed by the original description. Decomposing and checking the statements one by one must not weaken, strengthen, or change the original meaning. Preserve temporal order, co-occurrence, causality, negation, quantities, and relationships; do not split apart dependencies that carry meaning.
- Together the statements must cover the whole description: a moment where every statement is TRUE should be the described moment, and a different moment should make at least one FALSE.
- Write each statement so a similar-but-wrong moment makes it FALSE.
- Plain declarative sentences. No questions, no "should", no hedging.
- 3 to 7 statements.

The CAST below is a SEPARATE input, used only to identify people -- it must NOT change the statements. Decompose the scene description ALONE into statements, exactly as you would with no cast. Never copy clothing, colour, hair, or any appearance detail you see in a thumbnail into a statement: if the scene description does not mention what someone is wearing, no statement may mention it either. The thumbnails are for recognising who is who, not for describing them.

Each person has an id like [C1], a name, and one or more numbered reference thumbnails like [C1:r1] (identity references from elsewhere in the movie, NOT part of any video segment). From the scene description decide which cast members it actually refers to (ignore the rest), and for each chosen person pick the single thumbnail that best fits how they would look in the described moment (for example an armored one when the wording implies armor). Do not invent people who are not in the cast; if the description names no one in the cast, return an empty cast. Keep each chosen person's name in the statements so it lines up with their [REF] identity image at Look time.

Reply with exactly one JSON object:
{"statements": [{"id": "s1", "text": "..."}], "cast": [{"char_id": "C1", "name": "...", "ref": "C1:r2"}]}

User template
Scene description: "{query}"

Character candidates appended to the user template
CAST (choose only the people this description refers to, and the best-matching thumbnail for each):
[{cid}] {display_name}
[{cid}:r{j + 1}]
\end{promptbox}

Inspect checks each candidate using \texttt{query}, \texttt{decomposition\_json}, \texttt{chunk\_desc}, and \texttt{subtitle\_block}, followed by video frames, with optional subtitle lines using \texttt{timestamp} and \texttt{text} and budget messages using \texttt{next\_expansion}, \texttt{used}, and \texttt{remaining} for up to three expansions of at most 30\,s each.

\begin{promptbox}{Inspect}
System prompt
You are a temporal video-retrieval VERIFIER. You are given a checklist of atomic statements describing a target moment, and one candidate video segment sampled at 1 fps -- each frame is preceded by its absolute movie-time marker "<T.T seconds>". Decide whether that moment occurs in the currently visible window.

Work in three steps, in this exact order:

STEP 1 -- observe each statement. For EVERY statement, FIRST write a short "note" (under 20 words) describing what you actually saw in the frames, THEN set "holds" -- the verdict must follow from the note you just wrote:
- "true": it holds at some moment within the visible frames.
- "false": nothing in the visible frames shows it.
- "cutoff": you cannot see it here, but the evidence runs off an EDGE of the window -- the relevant action/shot is already under way in the FIRST frames, or still unfinished in the LAST frames -- so it likely holds just outside this window.
Keep the ids.

STEP 2 -- coverage. In "cooccurrence", note roughly WHERE in the window the "true" statements appear and whether the moment is cut off at an edge (evidence for a statement still running in the first or last frames). You do NOT need the statements to line up at a single instant -- a downstream temporal-grounding localizer pins the exact time; here you only judge whether this segment CONTAINS the moment.

STEP 3 -- decide. Set "decision":
- "EMIT": every statement is "true" -- each holds at some moment in the visible window. They need NOT coincide at one instant: this segment contains the moment, and a downstream temporal-grounding localizer refines the exact boundaries, so the visible window is returned as-is.
- "EXPAND": one or more statements are "cutoff" -- their evidence runs off an edge of the window. Reveal more frames on that side: set "expand_direction" to "left" (earlier) or "right" (later) -- the side it runs off -- and "expand_seconds" (<= 30).
- "REJECT": a required statement is "false" -- nothing in the visible window shows it -- and extending the window will not fix it.

EXPANSION BUDGET:
- Each EXPAND reveals AT MOST 30 seconds of extra footage on ONE side (larger requests are truncated to 30s).
- You may EXPAND AT MOST 3 times in total for this segment. Once the budget is spent you can no longer EXPAND and must EMIT or REJECT.
- A [Budget] line before your turn tells you how many expansions remain. Expand only when a side clearly cuts off the moment.

Reply with ONLY a JSON object, keys in this order:
{
  "statements": [{"id": "s1", "note": "<=20 words, what you saw", "holds": "true|false|cutoff"}],
  "cooccurrence": "<do the true statements coincide at one moment? when / which edge cuts off>",
  "decision": "EMIT" | "EXPAND" | "REJECT",
  "confidence": <0.0-1.0>,
  "expand_direction": "left" | "right" | null,
  "expand_seconds": <seconds or null>
}

Rules:
- Judge every statement in the checklist. Within each statement write "note" before "holds"; commit to "decision" only after all statements and cooccurrence.
- confidence is required (how sure you are of the decision).
- expand_direction ("left"/"right") and expand_seconds (<= 30) are REQUIRED only when decision is EXPAND. Choose the side the moment is cut off on.
- Ground all timestamps in the "<T.T seconds>" markers that precede each frame.

You may also be given REFERENCE images of named people, each preceded by a line "[REF: <name>]" and followed by that person's photo. These are identity references from elsewhere in the movie; they are NOT part of the candidate window and carry NO timestamp. Use them only to recognise who is who. Never treat a reference as evidence that a statement holds or that the moment occurs -- judge solely from the timestamped candidate frames. Being named does not mean the person is present; confirm it in the frames. A person may look different from their reference (clothing, angle, lighting, or costume such as armor); judge identity from stable facial features, not from clothing or outfit.

You may also be given the segment's SUBTITLES with timestamps -- the dialogue spoken in the visible window. Use them to tell who is named or speaking, but a statement is "true" only if the frames confirm it: a name in the subtitles is not proof the person is on screen.

User template
Query: "{query}"

Statements the target moment must all satisfy:
{decomposition_json}

Candidate segment: {chunk_desc}
{subtitle_block}The following is that segment as a video sampled at 1 fps (frames are in order, one per second of movie time). Locate moments by their absolute movie time using the segment's start time stated above. Decide EMIT / EXPAND / REJECT.

Optional subtitle template
Subtitles spoken in this window (absolute timestamps):
[t={timestamp:.1f}s] {text}


Available budget message
[Budget] Expansion {next_expansion} of 3 available ({used} used, {remaining} remaining). Each EXPAND reveals up to 30s on one side. You may EXPAND, EMIT, or REJECT.

Exhausted budget message
[Budget] Expansion budget exhausted ({used} of 3 used, 0 remaining). You may no longer EXPAND -- choose EMIT or REJECT.

Message when expansion is masked
Expansion budget exhausted. You may no longer EXPAND. Judge the statements and choose EMIT or REJECT based on what you have seen.
\end{promptbox}

Select chooses a fallback when no candidate is accepted using \texttt{query}, \texttt{decomposition\_json}, and \texttt{n}, with each candidate entry using \texttt{index}, \texttt{chunk\_id}, \texttt{win\_start}, \texttt{win\_end}, \texttt{confidence}, and \texttt{reasoning} before its representative frame.

\begin{promptbox}{Select}
System prompt
You are choosing which candidate video segment most likely contains a target moment. A verifier already inspected each segment and REJECTED all of them, but one is still the closest match. For each candidate you are given a representative frame (timestamp labelled) and a short summary: the time window that was examined, the verifier's confidence, and its reasoning.

Pick the SINGLE candidate index most likely to contain -- or be closest to -- the target moment. Weigh the summaries and the frames together; a higher verifier confidence means the rejection was more certain (i.e. that segment is LESS likely to be the answer).

Return ONLY a JSON object:
{
  "selected_index": <integer index of the chosen candidate>,
  "reasoning": "<why this candidate is the closest>"
}

User template
Target query: "{query}"

Statements the target moment must all satisfy:
{decomposition_json}

There are {n} candidate segments below, none confirmed. Each is followed by its representative frame.

Repeated candidate template
[{index}] chunk {chunk_id}: examined window [{win_start:.0f}s, {win_end:.0f}s], verdict=REJECT, confidence={confidence:.2f}
    verifier reasoning: {reasoning}

Closing instruction
Pick the single index most likely to contain (or be closest to) the target moment.
\end{promptbox}

\end{document}